\documentclass[10pt,twocolumn,letterpaper]{article}

\usepackage{iccv}
\usepackage{times}
\usepackage{epsfig}
\usepackage{graphicx}
\usepackage{amsmath}
\usepackage{amssymb}
\usepackage{multirow}
\usepackage{bbm}
\usepackage{stackengine}

\usepackage{array,etoolbox}
\newcounter{rownumbers}
\newcommand\rownumber{\stepcounter{rownumbers}\arabic{rownumbers}}

\DeclareMathOperator*{\argmax}{arg\,max}

\mathchardef\mhyphen="2D

\usepackage[pagebackref=true,breaklinks=true,letterpaper=true,colorlinks,bookmarks=false]{hyperref}
\usepackage{floatrow}
\DeclareFloatFont{tiny}{\tiny}
\DeclareFloatFont{small}{\small}
\DeclareFloatFont{footnotesize}{\footnotesize}
\usepackage[font=footnotesize,labelfont=bf]{caption}
\newcommand{\squishlist}{
 \begin{list}{$\bullet$}
  { \setlength{\itemsep}{0pt}
     \setlength{\parsep}{2pt}
     \setlength{\topsep}{2pt}
     \setlength{\partopsep}{0pt}
     \setlength{\leftmargin}{1.5em}
     \setlength{\labelwidth}{1em}
     \setlength{\labelsep}{0.5em} 
  } 
}
\newcommand{\squishend}{\end{list}}

\iccvfinalcopy 

\def\iccvPaperID{784} 

\ificcvfinal\fi
\begin{document}

\title{Phrase Localization and Visual Relationship Detection with Comprehensive Image-Language Cues}
\author{Bryan A. Plummer\\
\and
Arun Mallya\\
\and 
Christopher M. Cervantes\\
\and 
Julia Hockenmaier\\
\and 
Svetlana Lazebnik\\
University of Illinois at Urbana-Champaign \\
{\tt\small \{bplumme2, amallya2, ccervan2, juliahmr, slazebni\}@illinois.edu}
}

\maketitle
\thispagestyle{empty}

\begin{abstract}
This paper presents a framework for localization or grounding of phrases in images using a large collection of linguistic and visual cues. We model the appearance, size, and position of entity bounding boxes, adjectives that contain attribute information, and spatial relationships between pairs of entities connected by verbs or prepositions. Special attention is given to relationships between people and clothing or body part mentions, as they are useful for distinguishing individuals. We automatically learn weights for combining these cues and at test time, perform joint inference over all phrases in a caption. The resulting system produces state of the art performance on phrase localization on the Flickr30k Entities dataset~\cite{flickrentitiesijcv} and visual relationship detection on the Stanford VRD dataset~\cite{lu2016visual}.\footnote{Code:  \url{https://github.com/BryanPlummer/pl-clc}}
\end{abstract} \vspace{-10pt}

\section{Introduction}
Today's deep features can give reliable signals about a broad range of content in natural images, leading to advances in image-language tasks such as automatic captioning~\cite{fang2014captions,densecap2015,karpathy2014deep,karpathy2014deepNIPS,xu2015show} and visual question answering~\cite{AntolICCV2015,fukui16emnlp,yuICCV2015}. A basic building block for such tasks is localization or grounding of individual phrases~\cite{fang2014captions,karpathy2014deep,karpathy2014deepNIPS,ma2015,flickrentitiesijcv,wang2016CVPR,xu2015show}. A number of datasets with phrase grounding information have been released, including Flickr30k Entities~\cite{flickrentitiesijcv}, ReferIt~\cite{kazemzadeh-EtAl:2014:EMNLP2014}, Google Referring Expressions~\cite{mao2016generation}, and Visual Genome~\cite{krishnavisualgenome}.  However, grounding remains challenging due to open-ended vocabularies, highly unbalanced training data, prevalence of hard-to-localize entities like clothing and body parts, as well as the subtlety and variety of linguistic cues that can be used for localization.

\begin{figure}
\centering
\includegraphics[width=\textwidth]{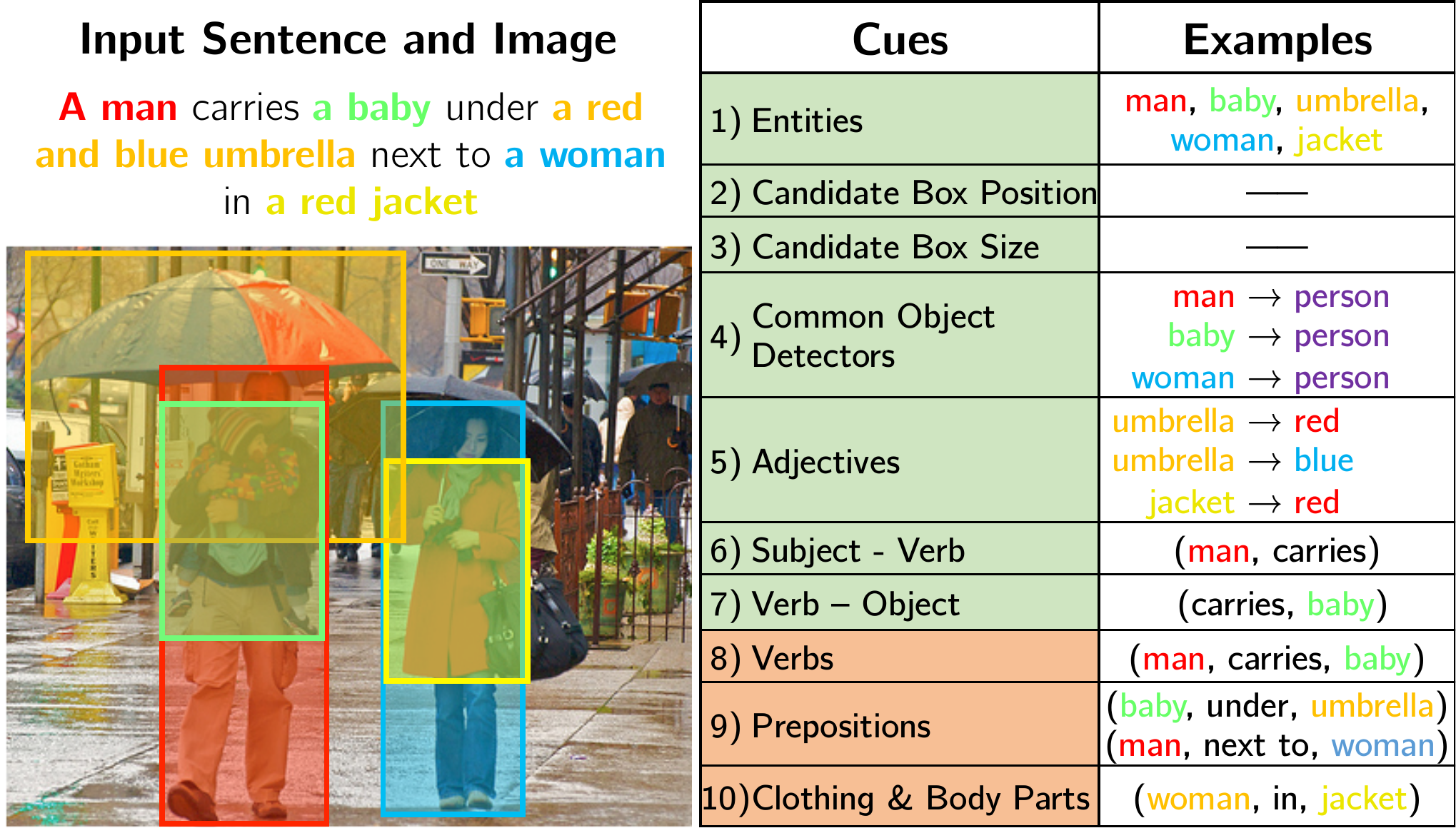}
\vspace{-20pt}
\caption{Left: an image and caption, together with ground truth bounding boxes of entities (noun phrases). Right: a list of all the cues used by our system, with corresponding phrases from the sentence.}

\label{fig:overviewFigure}
\end{figure}

\begin{table*}
\setlength{\tabcolsep}{3pt}
\centering
\small
\begin{tabular}{|ll|c|c|c|c|c|c|c|c|c|}
\hline
& \multirow{3}{*}{Method} & \multicolumn{6}{|c|}{Single Phrase Cues} & \multicolumn{2}{|c|}{Phrase-Pair Spatial Cues} & Inference\\
\cline{3-11}
& & Phrase-Region & Candidate & Candidate & Object & \multirow{2}{*}{Adjectives} & \multirow{2}{*}{Verbs} & Relative & Clothing \& & Joint\\
& & Compatibility & Position & Size & Detectors  &  &  & Position & Body Parts  & Localization\\
\hline
\hline
&Ours & \checkmark & \checkmark & \checkmark & \checkmark* & \checkmark & \checkmark & \checkmark & \checkmark & \checkmark\\
\hline
(a)&NonlinearSP~\cite{wang2016CVPR} & \checkmark & -- & -- & -- & -- & -- & --  & -- & --\\
&GroundeR~\cite{rohrbach2015} & \checkmark & -- & -- & -- & -- & -- & --  & -- & --\\
&MCB~\cite{fukui16emnlp} & \checkmark & -- & -- & -- & -- & -- & -- & -- & --\\
&SCRC~\cite{hu2015natural} & \checkmark & \checkmark & -- & -- & -- & -- & --  & -- & --\\
&SMPL~\cite{wang2016matching} & \checkmark & -- & -- & -- & -- & --  & \checkmark* & -- & \checkmark\\
&RtP~\cite{flickrentitiesijcv} & \checkmark & -- & \checkmark & \checkmark* & \checkmark* & -- & --  & -- & --\\
\hline
(b)
&Scene Graph~\cite{Johnson2015CVPR} & -- & -- & -- & \checkmark & \checkmark & -- & \checkmark & -- & \checkmark\\
&ReferIt~\cite{kazemzadeh-EtAl:2014:EMNLP2014} & -- & \checkmark & \checkmark & \checkmark & \checkmark* & -- & \checkmark  & -- & --\\
&Google RefExp~\cite{mao2016generation} & \checkmark & \checkmark & \checkmark & -- & -- & -- & --  & -- & --\\
\hline
\end{tabular}
\vspace{-2mm}
\caption{Comparison of cues for phrase-to-region grounding. {\bf (a)} Models applied to phrase localization on Flickr30K Entities. {\bf (b)} Models on related tasks.  * indicates that the cue is used in a limited fashion, i.e.\ ~\cite{kazemzadeh-EtAl:2014:EMNLP2014,flickrentitiesijcv} restricted their adjective cues to colors, ~\cite{wang2016matching} only modeled possessive pronoun phrase-pair spatial cues ignoring verb and prepositional phrases, ~\cite{flickrentitiesijcv} and we limit the object detectors to 20 common categories.\vspace{-17pt}}
\label{tab:cueComparison}
\end{table*}

The goal of this paper is to accurately localize a bounding box for each entity (noun phrase) mentioned in a caption for a particular test image.  We propose a joint localization objective for this task using a learned combination of single-phrase and phrase-pair cues. Evaluation is performed on the challenging recent Flickr30K Entities dataset~\cite{flickrentitiesijcv}, which provides ground truth bounding boxes for each entity in the five captions of the original Flickr30K dataset~\cite{young2014image}.  

Figure~\ref{fig:overviewFigure} introduces the components of our system using an example image and caption. Given a noun phrase extracted from the caption, e.g., {\em red and blue umbrella}, we obtain single-phrase cue scores for each candidate box based on appearance (modeled with a phrase-region embedding as well as object detectors for common classes), size, position, and attributes (adjectives). If a pair of entities is connected by a verb ({\em man carries a baby}) or a preposition ({\em woman in a red jacket}), we also score the pair of corresponding candidate boxes using a spatial model. In addition, actions may modify the appearance of either the subject or the object (e.g., a man carrying a baby has a characteristic appearance, as does a baby being carried). To account for this, we learn subject-verb and verb-object appearance models for the constituent entities. We give special treatment to relationships between people, clothing, and body parts, as these are commonly used for describing individuals, and are also among the hardest entities for existing approaches to localize. To extract as complete a set of relationships as possible, we use natural language processing (NLP) tools to resolve pronoun references within a sentence: e.g., by analyzing the sentence {\em A man puts his hand around a woman}, we can determine that the hand belongs to the man and introduce the respective pairwise term into our objective. 

Table~\ref{tab:cueComparison} compares the cues used in our work to those in other recent papers on phrase localization and related tasks like image retrieval and referring expression understanding. To date, other methods applied to the Flickr30K Entities dataset~\cite{fukui16emnlp,hu2015natural,rohrbach2015,wang2016CVPR,wang2016matching} have used a limited set of single-phrase cues. Information from the rest of the caption, like verbs and prepositions indicating spatial relationships, has been ignored. One exception is Wang~\etal~\cite{wang2016matching}, who tried to relate multiple phrases to each other, but limited their relationships only to those indicated by possessive pronouns, not personal ones. By contrast, we use pronoun cues to the full extent by performing pronominal coreference. Also, ours is the only work in this area incorporating the visual aspect of verbs. 
Our formulation is most similar to that of~\cite{flickrentitiesijcv}, but with a larger set of cues, learned combination weights, and a global optimization method for simultaneously localizing all the phrases in a sentence.  

In addition to our experiments on phrase localization, we also adapt our method to the recently introduced task of visual relationship detection (VRD) on the Stanford VRD dataset~\cite{lu2016visual}. Given a test image, the goal of VRD is to detect all entities and relationships present and output them in the form {\em (subject,
predicate, object)} with the corresponding bounding boxes. 
By contrast with phrase localization, where we are given a set of entities and
relationships that are in the image, in VRD we do not know {\em a priori} which objects or relationships might be present. On this task, our model shows significant performance gains over prior work, with especially acute differences in zero-shot detection due to modeling cues with a vision-language embedding.  This adaptability to never-before-seen examples is also a notable distinction between our approach and prior methods on related tasks (e.g.~\cite{fidler2013,Johnson2015CVPR,kazemzadeh-EtAl:2014:EMNLP2014,kong2014you}), which typically train their models on a set of predefined object categories, providing no support for out-of-vocabulary entities. 



Section \ref{sec:multi_cue_model} discusses our global objective function for simultaneously localizing all phrases from the sentence and describes the procedure for learning combination weights. Section~\ref{sec:extracting} details how we parse sentences to extract entities, relationships, and other relevant linguistic cues. Sections \ref{sec:singlePhraseCues} and \ref{sec:spatialCues} define single-phrase and phrase-pair cost functions between linguistic and visual cues.  Section~\ref{sec:expts} presents an in-depth evaluation of our cues on Flickr30K Entities~\cite{flickrentitiesijcv}. Lastly, Section~\ref{sec:expts_vrd} presents the adaptation of our method to the VRD task~\cite{lu2016visual}.

\section{Phrase localization approach}
\label{sec:multi_cue_model}

We follow the task definition used in~\cite{fukui16emnlp,hu2015natural,flickrentitiesijcv,rohrbach2015,wang2016CVPR,wang2016matching}: At test time, we are given an image and a caption with a set of entities (noun phrases), and we need to localize each entity with a bounding box. Section \ref{grounding_framework} describes our inference formulation, and Section \ref{sec:combiningWts} describes our procedure for learning the weights of different cues.

\subsection{Joint phrase localization}
\label{grounding_framework}
For each image-language cue derived from a single phrase or a pair of phrases (Figure \ref{fig:overviewFigure}), we define a \emph{cue-specific cost function} that measures its compatibility with an image region (small values indicate high compatibility). We will describe the cost functions in detail in Section \ref{sec:cues}; here, we give our test-time optimization framework for jointly localizing all phrases from a sentence.

Given a single phrase $p$ from a test sentence, we score each region (bounding box) proposal $b$ from the test image based on a linear combination of cue-specific cost functions $\phi_{\{1,\cdots,K_S\}}(p,b)$ with learned weights $w^S$:
\begin{equation}
\medmuskip=1mu
\thinmuskip=1mu
\thickmuskip=1mu
S(p,b;w^S) = \sum^{K_S}_{s=1}\mathbbm{1}_s(p) \phi_s(p,b) w^S_s,
\label{eq:unaryScore}
\end{equation}

\noindent where $\mathbbm{1}_s(p)$ is an indicator function for the availability of cue $s$ for phrase $p$ (e.g., an adjective cue would be available for the phrase \emph{blue socks}, but would be unavailable for \emph{socks} by itself). 
As will be described in Section \ref{sec:singlePhraseCues}, we use 14 single-phrase cost functions: region-phrase compatibility score, phrase position, phrase size (one for each of the eight phrase types of~\cite{flickrentitiesijcv}), object detector score, adjective, subject-verb, and verb-object scores.  

For a pair of phrases with some relationship $r=(p,rel,p')$ and candidate regions $b$ and $b'$, an analogous scoring function is given by a weighted combination of pairwise costs $\psi_{\{ 1,\cdots,K_Q \}}(r, b, b')$:
\begin{equation}
\medmuskip=1mu
\thinmuskip=1mu
\thickmuskip=1mu
Q(r,b,b';w^Q) = \sum^{K_Q}_{q=1}\mathbbm{1}_q(r) \psi_q(r,b,b') w^Q_q \,.
\label{eq:spatialScore}
\end{equation}
We use three pairwise cost functions corresponding to spatial classifiers for verb, preposition, and clothing and body parts relationships (Section \ref{sec:spatialCues}). 

We train all cue-specific cost functions on the training set and the combination weights on the validation set.
At test time, given an image and a list of phrases $\{p_1,\cdots,p_N\}$, we first retrieve top $M$ candidate boxes for each phrase $p_i$ using Eq.~(\ref{eq:unaryScore}). Our goal is then to select one bounding box $b_i$ out of the $M$ candidates per each phrase $p_i$ such that the following objective is minimized:
\begin{equation}
\medmuskip=0mu
\thinmuskip=0mu
\thickmuskip=0mu
\min_{b_1,\cdots,b_N} \left\{ \sum_{p_i} S(p_i, b_i)\quad + \kern-0.5em \sum_{r_{ij} = (p_i, rel_{ij}, p_j)} \kern-1.0em Q(r_{ij}, b_i, b_j) \right\}
\label{eq:energy}
\end{equation}
where phrases $p_i$ and $p_j$ (and respective boxes $b_i$ and $b_j$) are related by some relationship $rel_{ij}$. This is a binary quadratic programming formulation inspired by~\cite{tighe2014scene}; we relax and solve it using a sequential QP solver in MATLAB.
The solution gives a single bounding box hypothesis for each phrase. Performance is evaluated using Recall@1, or proportion of phrases where the selected box has Intersection-over-Union (IOU) $\geq 0.5$ with the ground truth.

\subsection{Learning scoring function weights}
\label{sec:combiningWts}

We learn the weights $w^S$ and $w^Q$ in Eqs. (\ref{eq:unaryScore}) and (\ref{eq:spatialScore}) by directly optimizing recall on the validation set. 
We start by finding the unary weights $w^S$ that maximize the number of correctly localized phrases:
\begin{equation}
w^S = \argmax_{w} \sum^N_{i=1}\mathbbm{1}_{IOU \geq 0.5}(b^*_i, \hat{b}(p_i; w)),
\label{eq:unary_objective}
\end{equation}

\noindent where $N$ is the number of phrases in the training set, 
$\mathbbm{1}_{IOU \geq 0.5}$ is an indicator function returning 1 if the two boxes have IOU $\ge$ 0.5,
$b^*_i$ is the ground truth bounding box for phrase $p_i$,
$\hat{b}(p;w)$ returns the most likely box candidate for phrase $p$ under 
the current weights, or, more formally, given a set of candidate boxes 
$\mathcal{B}$,
\begin{equation}
\hat{b}(p; w) = \min_{b \in \mathcal{B}} S(p,b;w).
\label{eq:unaryBoxSelection}
\end{equation}
\noindent We optimize Eq.\ (\ref{eq:unary_objective}) using a derivative-free direct search method~\cite{lagarias98} (MATLAB's fminsearch). We randomly initialize the weights, keep the best weights after 20 runs based on validation set performance (takes just a few minutes to learn weights for all single phrase cues in our experiments).

Next, we fix $w^S$ and learn the weights $w^Q$ over phrase-pair cues in the validation set. To this end, we formulate an objective analogous to Eq.~(\ref{eq:unary_objective}) for maximizing the number of correctly localized region pairs. Similar to Eq.~(\ref{eq:unaryBoxSelection}), we define the function $\hat{\rho}(r;w)$ to return the best pair of boxes for the relationship $r = (p, rel, p')$:
\begin{equation}
\medmuskip=0mu
\thinmuskip=0mu
\thickmuskip=0mu
\hat{\rho}(r; w) = \min_{b,b' \in \mathcal{B}} S(p,b;w^S) + S(p',b';w^S) + Q(r,b,b';w).
\label{eq:boxSelection}
\end{equation}
Then our pairwise objective function is
\begin{equation}
w^Q = \argmax_{w} \sum^M_{k=1}\mathbbm{I}_{Pair IOU \geq 0.5}(\rho^*_k,\hat{\rho}(r_k; w)),
\label{eq:pairwise_objective}
\end{equation}

\noindent where $M$ is the number of phrase pairs with a relationship, $\mathbbm{I}_{Pair IOU \geq 0.5}$ returns the number of correctly localized boxes (0, 1, or 2), and $\rho^*_k$ is the ground truth box pair for the relationship $r_k = (p_k, rel_k, p_k')$. 

Note that we also attempted to learn the weights $w^S$ and $w^Q$ using standard approaches such as rank-SVM~\cite{joachims2006training}, but found our proposed direct search formulation to work better. In phrase localization, due to its Recall@1 evaluation criterion, only the correctness of one best-scoring candidate region for each phrase matters, unlike in typical detection scenarios, where one would like all positive examples to have better scores than all negative examples. The VRD task of Section \ref{sec:expts_vrd} is a more conventional detection task, so there we found rank-SVM to be more appropriate.

\section{Cues for phrase-region grounding}
\label{sec:cues}

Section \ref{sec:extracting} describes how we extract linguistic cues from sentences. Sections \ref{sec:singlePhraseCues} and \ref{sec:spatialCues} give our definitions of the two types of cost functions used in Eqs. (\ref{eq:unaryScore}) and (\ref{eq:spatialScore}): single phrase cues (SPC) measure the compatibility of a given phrase with a candidate bounding box, and phrase pair cues (PPC) ensure that pairs of related phrases are localized in a spatially coherent manner. 

\subsection{Extracting linguistic cues from captions} \label{sec:extracting}

The Flickr30k Entities dataset provides annotations for Noun Phrase (NP) chunks corresponding to entities, but linguistic cues corresponding to adjectives, verbs, and prepositions must be extracted from the captions using NLP tools. Once these cues are extracted, they will be translated into visually relevant constraints for grounding. In particular, we will learn specialized detectors for adjectives, subject-verb, and verb-object relationships (Section~\ref{detectors}). Also, because pairs of entities connected by a verb or preposition have constrained layout, we will train classifiers to score pairs of boxes based on spatial information (Section~\ref{sec:spatialCues}). 

Adjectives are part of NP chunks so identifying them is trivial. To extract other cues, such as verbs and prepositions that may indicate actions and spatial relationships, we obtain a constituent parse tree for each sentence using the Stanford parser~\cite{SocherEtAl2013:CVG}. 
Then, for possible relational phrases (prepositional and verb phrases), 
we use the method of Fidler~\etal~\cite{fidler2013}, where we start at the 
relational phrase and then traverse up the tree and to the left until we reach a noun phrase node, which will correspond to the first entity in an \emph{(entity1, rel, entity2)} tuple. The second entity is given by the first noun phrase node on the right side of the relational phrase in the parse tree. For example, given the sentence \emph{A boy running in a field with a dog}, the extracted NP chunks would be \emph{a boy, a field, a dog}. The relational phrases would be \emph{(a boy, running in, a field)} and \emph{(a boy, with, a dog)}.

Notice that a single relational phrase can give rise to multiple relationship cues. Thus, from {\em (a boy, running in, a field)}, we extract the verb relation {\em (boy, running, field)} and prepositional relation {\em (boy, in, field)}. An exception to this is a relational phrase where the first entity is a person and the second one is of the clothing or body part type,\footnote{Each NP chunk from the Flickr30K dataset is classified into one of eight phrase types based on the dictionaries of~\cite{flickrentitiesijcv}.} e.g., {\em (a boy, running in, a jacket)}. For this case, we create a single special pairwise relation {\em (boy, jacket)} that assumes that the second entity is attached to the first one and the exact relationship words do not matter, i.e., {\em (a boy, running in, a jacket)} and {\em (a boy, wearing, a jacket)} are considered to be the same. The attachment assumption can fail for phrases like {\em (a boy, looking at, a jacket)}, but such cases are rare. 


Finally, since pronouns in Flickr30k Entities are not annotated, we attempt to perform pronominal coreference (i.e., creating a link between a pronoun and the phrase it refers to) in order to extract a more complete set of cues. As an example, given the sentence {\em Ducks feed themselves}, initially we can only extract the subject-verb cue $(ducks, feed)$, but we don't know who or what they are feeding. Pronominal coreference resolution tells us that the ducks are themselves eating and not, say, feeding ducklings.  
We use a simple rule-based method similar to knowledge-poor methods~\cite{harabagiu1999knowledge,mitkov1998robust}.
Given lists of pronouns by type,\footnote{Relevant pronoun types are subject, object, reflexive, reciprocal, relative, and indefinite.} our rules attach each pronoun with at most one non-pronominal mention that occurs earlier in the sentence (an antecedent). 
We assume that subject and object pronouns often refer to the main subject (e.g.\ \emph{[A dog] laying on the ground looks up at the dog standing over [him]}), reflexive and reciprocal pronouns refer to the nearest antecedent (e.g.\ \emph{[A tennis player] readies [herself]}.), and indefinite pronouns do not refer to a previously described entity. It must be noted that compared with verb and prepositional relationships, relatively few additional cues are extracted using this procedure (432 pronoun relationships in the test set and 13,163 in the train set, while the counts for the other relationships are on the order of 10K and 300K).

\subsection{Single Phrase Cues (SPCs)}
\label{sec:singlePhraseCues}
\noindent{\bf Region-phrase compatibility:}  This is the most basic cue relating
phrases to image regions based on appearance. It is applied to every test phrase (i.e., its indicator function in Eq.\ (\ref{eq:unaryScore}) is always 1). 
Given phrase $p$ and region $b$, the cost $\phi_{CCA}(p,b)$ is given by the cosine distance between $p$ and $b$ in a joint embedding space learned using normalized Canonical Correlation Analysis (CCA)~\cite{gong14}. We use the same procedure as~\cite{flickrentitiesijcv}. 
Regions are represented by the fc7 activations of a Fast-RCNN model~\cite{girshickICCV15fastrcnn} fine-tuned using the union of the PASCAL 2007 and 2012 trainval sets~\cite{pascal-voc-2012}.  After removing stopwords, phrases are represented by the HGLMM fisher vector encoding~\cite{klein2014fisher} of word2vec~\cite{mikolov2013efficient}.  


\smallskip
\noindent{\bf Candidate position:}  The location of a bounding box in an image has 
been shown to be predictive of the kinds of phrases it may refer 
to~\cite{divvalaCVPR2009,hu2015natural,kazemzadeh-EtAl:2014:EMNLP2014,liIJCV2014}. 
We learn location models for each of the eight broad phrase types specified in~\cite{flickrentitiesijcv}: people, clothing, body parts, vehicles, animals, scenes, and a catch-all ``other.'' We represent a bounding box by its centroid normalized by the image size, the percentage of the image covered by the box, and its aspect ratio, 
resulting in a 4-dim.\ feature vector. We then train a support vector machine (SVM) with a radial basis function (RBF) kernel using LIBSVM~\cite{CC01a}.  We randomly sample EdgeBox~\cite{ZitnickECCV14} proposals with $\text{IOU} < 0.5$ with the ground truth boxes for negative examples. Our scoring function is
\begin{equation*}
\phi_{pos}(p, b) = - \log(\text{SVM}_{type(p)}(b)),
\end{equation*}
\noindent where SVM$_{type(p)}$ returns the probability that box 
$b$ is of the phrase type $type(p)$ (we use Platt scaling~\cite{platt99} to convert the SVM output to a probability). 

\smallskip
\noindent{\bf Candidate size:} People have a bias towards describing larger, more 
salient objects, leading prior work to consider the size of a candidate box in their models~\cite{fidler2013,kazemzadeh-EtAl:2014:EMNLP2014,flickrentitiesijcv}. 
We follow the procedure of~\cite{flickrentitiesijcv}, so that given a box $b$ with dimensions normalized by the image size, we have
\begin{equation*}
\phi_{size_{type(p)}}(p, b) = 1 - b_{width} \times b_{height}.
\end{equation*}

\noindent 
Unlike phrase position, this cost function does not use a trained SVM per phrase type. Instead, each phrase type is its own feature and the corresponding indicator function returns 1 if that phrase belongs to the associated type.
\smallskip

\noindent{\bf Detectors:}
\label{detectors}
CCA embeddings are limited in their ability to localize objects because they must account for a wide range of phrases and because they do not use negative examples during training. To  compensate for this, we use Fast R-CNN~\cite{girshickICCV15fastrcnn} to learn three networks for common object categories, attributes, and actions. Once a detector is trained, its score for a region proposal $b$ is

\begin{equation*}
\phi_{det}(p, b) = - \log(\text{softmax}_{det}(p, b)),
\end{equation*}

\noindent where softmax$_{det}(p, b)$ returns the output of the softmax layer 
for the object class corresponding to $p$. We manually create dictionaries to map phrases to detector categories (e.g., man, woman, \emph{etc.} map to `person'), and the indicator function for each detector returns 1 only if one of the words in the phrase exists in its dictionary. If multiple detectors for a single cue type are appropriate for a phrase (e.g., {\em a black and white shirt} would have two adjective detectors fire, one for each color), the scores are averaged. 
Below, we describe the three detector networks used in our model. Complete dictionaries can be found in Appendix~\ref{app:detector}.
\smallskip

\noindent{\bf Objects:}
We use the dictionary of~\cite{flickrentitiesijcv} to map nouns to 
the 20 PASCAL object categories~\cite{pascal-voc-2012} and fine-tune the network
on the union of the PASCAL VOC 2007 and 2012 trainval sets. At test time, when we run a detector for a phrase that maps to one of these object categories, we also use bounding box regression to refine the original region proposals. Regression is not used for the other networks below.
\smallskip

\noindent{\bf Adjectives:}  Adjectives found in phrases, especially color, 
provide valuable attribute information for localization~\cite{fidler2013,Johnson2015CVPR,kazemzadeh-EtAl:2014:EMNLP2014,flickrentitiesijcv}. 
The Flickr30K Entities baseline approach~\cite{flickrentitiesijcv} used a network trained for 11 colors. As a generalization of that, we create a list of adjectives that occur at least 100 times in the training set of Flickr30k.  After grouping together similar words and filtering out non-visual terms (e.g., \emph{adventurous}), we are left with a dictionary of 83 adjectives. As in~\cite{flickrentitiesijcv}, we consider color terms describing people ({\em black man}, {\em white girl}) to be separate categories. 
\smallskip

\noindent{\bf Subject-Verb and Verb-Object:} Verbs can modify the appearance of both the subject and the object in a relation. For example, knowing that a person is riding a horse can give us better appearance models for finding both the person and the horse~\cite{sadeghi2015viske,sadeghi2011vphrases}.  
As we did with adjectives, we collect verbs that occur at least 100 times in the training set, group together similar words, and filter out those that don't have a clear visual aspect, resulting in a dictionary of 58 verbs.
Since a person running looks different than a dog running, we subdivide our verb categories by phrase type of the subject (resp. object) if that phrase type occurs with the verb at least 30 times in the train set. For example, if there are enough animal-running occurrences, we create a new category with instances of all animals running. For the remaining phrases, we train a catch-all detector over all the phrases related to that verb. Following~\cite{sadeghi2015viske}, we train separate detectors for subject-verb and verb-object relationships, resulting in dictionary sizes of 191 (resp. 225). We also attempted to learn subject-verb-object detectors as in~\cite{sadeghi2015viske,sadeghi2011vphrases}, but did not see a further improvement.

\subsection{Phrase-Pair Cues (PPCs)}
\label{sec:spatialCues}
So far, we have discussed cues pertaining to a single phrase, but relationships between pairs of phrases can also provide cues about their relative position. 
We denote such relationships as tuples 
$(p_{\mathit{left}}, rel, p_{\mathit{right}})$ with $\mathit{left}, \mathit{right}$ indicating on which side of the relationship the phrases occur. As discussed in Section \ref{sec:extracting}, we consider three distinct types of relationships: verbs (\emph{man, riding, horse}), prepositions (\emph{man, on, horse}), and clothing and body parts (\emph{man, wearing, hat}). For each of the three relationship types, we group phrases referring to people but treat all other phrases as distinct, and then gather all relationships that occur at least 30 times in the training set. Then we learn a spatial relationship model as follows. 
Given a pair of boxes with coordinates $b = (x, y, w, h)$ and $b' = (x',y',w',h')$, we compute a four-dim.\ feature 
\begin{equation} \label{eq:spatial_feature}
\left[ (x - x')/w,\text{ }(y - y')/h,\text{ }w'/w,\text{ }h'/h \right],
\end{equation}

\noindent and concatenate it with combined SPC scores $S(p_{\mathit{left}},b)$, $S(p_{\mathit{right}},b')$ from Eq.~(\ref{eq:unaryScore}). To obtain negative examples, we randomly sample from other box pairings with $\text{IOU}<0.5$ with the ground truth regions from that image. We train an RBF SVM classifier with Platt scaling~\cite{platt99} to obtain a probability output.  This is similar to the method of~\cite{Johnson2015CVPR}, but rather than learning a Gaussian Mixture Model using only positive data, we learn a more discriminative model. Below are details on the three types of relationship classifiers.  Complete dictionaries can be found in Appendix~\ref{app:spatial}.
\smallskip

\noindent{\bf Verbs:} Starting with our dictionary of 58 verb detectors and following the above procedure of identifying all relationships that occur at least 30 times in the training set, we end up with 260 $(p_\mathit{left}, rel_\mathit{verb}, p_\mathit{right})$ SVM classifiers.
\smallskip

\noindent{\bf Prepositions:} 
We first gather a list of prepositions that occur at least 100 times in the training set, combine similar words, and filter out words that do not indicate a clear spatial relationship. This yields eight prepositions (\emph{in, on, under, behind, across, between, onto, and near}) and 216 $(p_\mathit{left}, rel_\mathit{prep}, p_\mathit{right})$ relationships.
\smallskip

\noindent{\bf Clothing and body part attachment:}  
We collect $(p_\mathit{left}, rel_\mathit{c\&bp}, p_\mathit{right})$ relationships 
where the left phrase is always a person and the right phrase is from the clothing or body part type and learn 207 such classifiers. As discussed in Section \ref{sec:extracting}, this relationship type takes precedence over any verb or preposition relationships that may also hold between the same phrases.

\section{Experiments on Flickr30k Entities} 
\label{sec:expts}

\subsection{Implementation details}

We utilize the provided train/test/val split of 29,873 training, 
1,000 validation, and 1,000 testing images~\cite{flickrentitiesijcv}. 
Following~\cite{flickrentitiesijcv}, our region proposals are given by the top 
200 EdgeBox~\cite{ZitnickECCV14} proposals per image.  
At test time, given a sentence and an image, we first use Eq.\ (\ref{eq:unaryScore}) to find the top 30 candidate regions for 
each phrase after performing non-maximum suppression using a 0.8 IOU threshold.
Restricted to these candidates, we optimize Eq.\ (\ref{eq:spatialScore}) to find a globally consistent mapping
of phrases to regions.


Consistent with~\cite{flickrentitiesijcv}, we only evaluate localization for phrases with a ground truth 
bounding box. If multiple bounding boxes are associated with a  phrase (e.g., four individual boxes for \emph{four men}), we represent the phrase as the union of its boxes. For each image and phrase in the test set, the predicted box must have at least 0.5 IOU with its ground truth box to be deemed successfully localized.  As only a single candidate is selected for each phrase, we report the proportion of correctly localized phrases (i.e.\ Recall@1).

\subsection{Results}

\begin{table}
    \begin{tabular}{|ll|c|}
      \hline
      & Method & Accuracy\\
      \hline
      \hline
      (a) & {\bf Single-phrase cues} & \\
      & CCA & 43.09 \\
      & CCA+Det & 45.29 \\
      & CCA+Det+Size & 51.45 \\
      & CCA+Det+Size+Adj & 52.63 \\
      & CCA+Det+Size+Adj+Verbs & 54.51 \\
      & CCA+Det+Size+Adj+Verbs+Pos (SPC) & {\bf 55.49 }\\
      \hline
      (b) & {\bf Phrase pair cues} & \\
      & SPC+Verbs & 55.53 \\
      & SPC+Verbs+Preps & 55.62 \\
      & SPC+Verbs+Preps+C\&BP (SPC+PPC) & {\bf 55.85}\\
      \hline
       (c) & {\bf State of the art} & \\
      & SMPL~\cite{wang2016matching} &  42.08 \\
      & NonlinearSP~\cite{wang2016CVPR} &  43.89 \\
      & GroundeR~\cite{rohrbach2015} &  47.81 \\
      & MCB~\cite{fukui16emnlp} &  48.69 \\
      & RtP~\cite{flickrentitiesijcv} &  50.89 \\
      \hline
    \end{tabular}
\vspace{-2mm}
\caption{Phrase-region grounding performance on the Flickr30k Entities dataset.  {\bf (a)} Performance of our single-phrase cues (Sec.~\ref{sec:singlePhraseCues}).  {\bf (b)} Further improvements by adding our pairwise cues (Sec.~\ref{sec:spatialCues}). {\bf (c)} Accuracies of competing state-of-the-art methods. This comparison excludes concurrent work that was published after our initial submission~\cite{ChenICMR2017}.\vspace{-1mm}}
\label{tab:overall_results}
\end{table}

Table~\ref{tab:overall_results} reports our overall localization accuracy for combinations of cues and compares our performance to the state of the art. Object detectors, reported on the second line of Table~\ref{tab:overall_results}(a), show a 2\% overall gain over the CCA baseline. This includes the gain from the detector score as well as the bounding box regressor trained with the detector in the Fast R-CNN framework~\cite{girshickICCV15fastrcnn}. Adding adjective, verb, and size cues improves accuracy by a further 9\%. Our last cue in Table~\ref{tab:overall_results}(a), position, provides an additional 1\% improvement.

We can see from Table~\ref{tab:overall_results}(b) that the spatial cues give only a small overall boost in accuracy on the test set, 
but that is due to the relatively small number of phrases to which they apply. In Table \ref{tab:sc_results} we will show that the localization improvement on the affected phrases is much larger. 

Table~\ref{tab:overall_results}(c) compares our performance to the state of the art. The method most similar to ours is our earlier model~\cite{flickrentitiesijcv}, which we call RtP here. RtP relies on a subset of our single-phrase cues (region-phrase CCA, size, object detectors, and color adjectives), and localizes each phrase separately. The closest version of our current model to RtP is CCA+Det+Size+Adj, which replaces the 11 colors of~\cite{flickrentitiesijcv} with our more general model for 83 adjectives, and obtains almost 2\% better performance. Our full model is 5\% better than RtP. It is also worth noting that a rank-SVM model~\cite{joachims2006training} for learning cue combination weights gave us 8\% worse performance than the direct search scheme of Section \ref{sec:combiningWts}.

Table~\ref{tab:type_results} breaks down the comparison by phrase type. Our model has the highest accuracy on most phrase types, with scenes being the most notable exception, for which GroundeR~\cite{rohrbach2015} does better. However, GroundeR uses Selective Search proposals~\cite{Uijlings2013SS}, which have an upper bound performance that is 7\% higher on scene phrases despite using half as many proposals. Although body parts have the lowest localization accuracy at 25.24\%, this represents an 8\% improvement in accuracy over prior methods.  However, only around 62\% of body part phrases have a box with high enough IOU with the ground truth, showing a major area of weakness of category-independent proposal methods.
Indeed, if we were to augment our EdgeBox region proposals with ground truth boxes, we would get an overall improvement in accuracy of about 9\% for the full system.

\begin{table*}
    \begin{tabular}{|l|c|c|c|c|c|c|c|c|}
      \hline
      & People &  Clothing & Body Parts & Animals & Vehicles & Instruments & Scene & Other \\
      \hline
      \hline
      \#Test & 5,656  & 2,306 & 523 & 518 & 400 & 162 & 1,619 & 3,374 \\
      \hline \hline
      SMPL~\cite{wang2016matching} & 57.89 & 34.61 & 15.87 & 55.98 & 52.25 & 23.46 & 34.22 & 26.23 \\
      GroundeR~\cite{rohrbach2015} & 61.00 & 38.12 & 10.33 & 62.55 & {\bf 68.75} & 36.42 & {\bf 58.18} & 29.08 \\
      RtP~\cite{flickrentitiesijcv} & 64.73 & 46.88 & 17.21 & 65.83 & {\bf 68.75} & {\bf 37.65} & 51.39 & 31.77 \\
      SPC+PPC (ours) & {\bf 71.69} & {\bf 50.95} & {\bf 25.24} & {\bf 76.25} & 66.50 & 35.80 & 51.51 & {\bf 35.98} \\
      \hline
      Upper Bound & 97.72 & 83.13 & 61.57 & 91.89 & 94.00 & 82.10 & 84.37 & 81.06 \\
      \hline
    \end{tabular}
    \vspace{-2mm}
    \caption{Comparison of phrase localization performance over phrase types. Upper Bound refers to the proportion of phrases of each type for which there exists a region proposal having at least $0.5$ IOU with the ground truth.  \vspace{-3mm}}
  \label{tab:type_results}
\end{table*}

\begin{table*} 
\centering
 \begin{tabular}{|l|r|r|r|r|r|r|r|r|r|r|}
  \hline
  & \multicolumn{4}{|c|}{Single Phrase Cues (SPC)} & \multicolumn{6}{|c|}{Phrase-Pair Cues (PPC)}\\ 
  \cline{2-11}
  \multirow{2}{*}{Method} & \multirow{3}{*}{\begin{tabular}{c}Object \\ Detectors\end{tabular}} & \multirow{3}{*}{Adjectives} & \multirow{3}{*}{\begin{tabular}{c}Subject- \\ Verb\end{tabular}} & \multirow{3}{*}{\begin{tabular}{c}Verb- \\ Object\end{tabular}} & \multicolumn{2}{|c|}{\multirow{2}{*}{Verbs}}  &  \multicolumn{2}{|c|}{\multirow{2}{*}{Prepositions}}  & \multicolumn{2}{|c|}{Clothing \&}\\
  &  &  &  & & \multicolumn{2}{|c|}{} & \multicolumn{2}{|c|}{} & \multicolumn{2}{|c|}{Body Parts}\\ 
  \cline{6-11}
  & & & & & \multicolumn{1}{|c|}{Left} & \multicolumn{1}{|c|}{Right} & \multicolumn{1}{|c|}{Left} & \multicolumn{1}{|c|}{Right} & \multicolumn{1}{|c|}{Left} & \multicolumn{1}{|c|}{Right}\\ \hline\hline
  Baseline & 74.25 & 57.71 & 69.68 & 40.70 & 78.32 & 51.05 & 68.97 & 55.01 & 81.01 & 50.72\\
  +Cue & 75.78 & 64.35 & 75.53 & 47.62 & 78.94 & 51.33 & 69.74 & 56.14 & 82.86 & 52.23\\
  \hline\hline
  \#Test & 4,059 & 3,809 & 3,094 & 2,398 & 867 & 858 & 780 & 778 & 1,464 & 1,591\\
  \hline
  \#Train & 114,748 & 110,415 & 94,353 & 71,336 & 26,254 & 25,898 & 23,973 & 23,903 & 42,084 & 45,496\\
  \hline
  \end{tabular}
  \vspace{-2mm}
\caption{Breakdown of performance for individual cues restricted only to test phrases to which they apply. For SPC, Baseline is given by CCA+Position+Size. For PPC, Baseline is the full SPC model.  For all comparisons, we use the improved boxes from bounding box regression on top of object detector output. PPC evaluation is split by which side of the relationship the phrases occur on. The bottom two rows show the numbers of affected phrases in the test and training sets. For reference, there are 14.5k visual phrases in the test set and 427k visual phrases in the train set.\vspace{-5mm}}
\label{tab:sc_results}
\end{table*}

\begin{figure*}
\centering
  \includegraphics[width=\textwidth] 
  {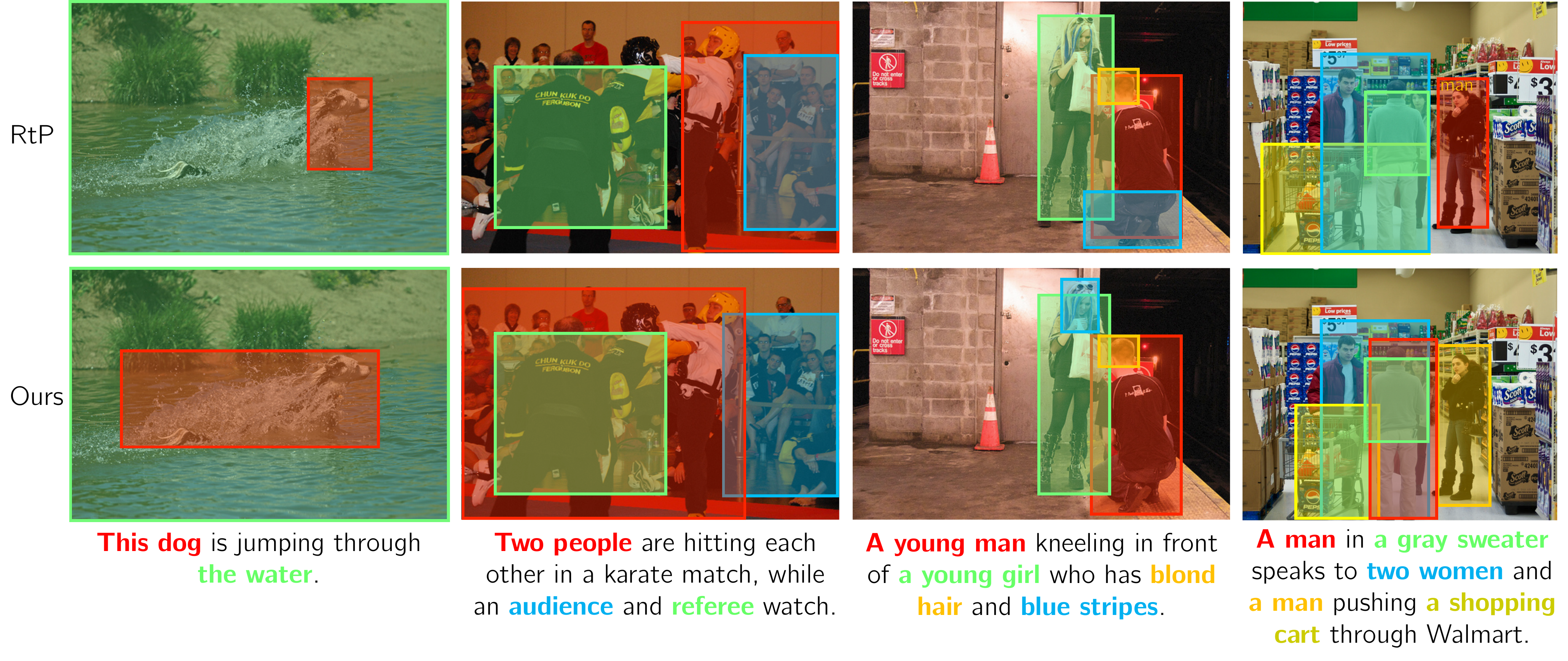}
  \vspace{-8mm}
  \caption{Example results on Flickr30k Entities comparing our SPC+PPC model's output with the RtP model~\cite{flickrentitiesijcv}. See text for discussion.\vspace{-4mm}}
  \label{fig:loc_examples}
\end{figure*}

Since many of the cues apply to a small subset of the phrases, 
Table~\ref{tab:sc_results} details the performance of cues over only the phrases they affect.
As a baseline, we compare against the combination of cues available for all phrases: region-phrase CCA, position, and size. To have a consistent set of regions, the baseline also uses improved boxes from bounding box regressors trained along with the object detectors.
As a result, the object detectors provide less than 2\% gain over the baseline for the phrases on which they are used,
suggesting that the regression provides the majority of the gain from CCA to CCA+Det in Table \ref{tab:overall_results}.
This also confirms that there is significant room for improvement in selecting candidate regions.  
By contrast, adjective, subject-verb, and verb-object detectors show 
significant gains, improving over the baseline by 6-7\%.

The right side of Table~\ref{tab:sc_results} shows the improvement on phrases due to phrase pair cues. Here, we separate the phrases that 
occur on the left side of the relationship, which corresponds to the subject, 
from the phrases on the right side. Our results 
show that the subject, is generally easier to localize. On the other hand, clothing and body parts show up mainly on the right side of relationships and they tend to be small. It is also less likely that such phrases will have good candidate boxes -- recall from Table~\ref{tab:type_results} that body parts have a performance upper bound of only 62\%. Although they affect relatively few test phrases, all three of our relationship classifiers show consistent gains over the SPC model. This is encouraging given that many of the relationships that are used on the validation set to learn our model parameters do not occur in the test set (and vice versa).

Figure~\ref{fig:loc_examples} provides a qualitative comparison of our output with the RtP model~\cite{flickrentitiesijcv}. 
In the first example, the prediction for the dog is improved due to the subject-verb classifier for \emph{dog jumping}. 
For the second example, pronominal coreference resolution (Section \ref{sec:extracting}) links {\em each other} to {\em two men}, telling us that not only is a man hitting something, but also that another man is being hit.  In the third example, the RtP model is not able to locate the woman's blue stripes in her hair despite having a model for {\em blue}. 
Our adjective detectors take into account \emph{stripes} as well as \emph{blue}, allowing us to correctly localize the phrase, even though we still fail to localize the hair.  Since the blue stripes and hair should co-locate, a method for obtaining co-referent entities would further improve performance on such cases. In the last 
example, the RtP model makes the same incorrect prediction for the two men. 
However, our spatial relationship between the first man and his gray sweater helps us correctly localize him. We also improve our prediction for the shopping cart.

\section{Visual Relationship Detection}
\label{sec:expts_vrd}
In this section, we adapt our framework to the recently introduced Visual Relationship Detection (VRD) benchmark of Lu~\etal~\cite{lu2016visual}. Given a test image without any text annotations, the task of VRD is to detect all entities and relationships present and output them in the form (\emph{subject, predicate, object}) with the corresponding bounding boxes. A relationship detection is judged to be correct if it exists in the image and both the subject and object boxes have IOU $\ge$ 0.5 with their respective ground truth. In contrast to phrase grounding, where we are given a set of entities and relationships that are assumed to be in the image, here we do not know {\em a priori} which objects or relationships might be present. On the other hand, the VRD dataset is easier than Flickr30K Entities in that it has a limited vocabulary of 100 object classes and 70 predicates annotated in 4000 training and 1000 test images.

Given the small fixed class vocabulary, it would seem advantageous to train 100 object detectors on this dataset, as was done by Lu~\etal~\cite{lu2016visual}. However, the training set is relatively small, the class distribution is unbalanced, and there is no validation set. Thus, we found that training detectors and then relationship models on the same images causes overfitting because the detector scores on the training images are overconfident. We obtain better results by training all appearance models using CCA, which also takes into account semantic similarity between category names and is trivially extendable to previously unseen categories.  Here, we use fc7 features from a Fast RCNN model trained on MSCOCO~\cite{lin2014microsoft} due to the larger range of categories than PASCAL, and word2vec for object and predicate class names. We train the following CCA models:
\vspace{-2mm}
\begin{enumerate} \setlength{\itemsep}{-1ex}
\item CCA(entity box, entity class name): this is the equivalent to region-phrase CCA in Section \ref{sec:singlePhraseCues} and is used to score both candidate subject and object boxes.
\item CCA(subject box, [subject class name, predicate class name]): analogous to subject-verb classifiers of Section \ref{sec:singlePhraseCues}. The 300-dimensional word2vec features of subject and predicate class names are concatenated.
\item CCA(object box, [predicate class name, object class name]): analogous to verb-object classifiers of Section \ref{sec:singlePhraseCues}.
\item CCA(union box, predicate class name): this model measures the compatibility between the bounding box of both subject and object and the predicate name. 
\item CCA(union box, [subject class name, predicate class name, object class name]).
\end{enumerate}
\vspace{-2mm}
Note that models 4 and 5 had no analogue in our phrase localization system. On that task, entities were known to be in the image and relationships simply provided constraints, while here we need to predict which relationships exist. To make predictions for predicates and relationships (which is the goal of models 4 and 5), it helps to see both the subject and object regions. Union box features were also less useful for phrase localization due to the larger vocabularies and relative scarcity of relationships in that task.

Each candidate relationship gets six CCA scores (model 1 above is applied both to the subject and the object). In addition, we compute size and position scores as in Section \ref{sec:singlePhraseCues} for subject and object, and a score for a pairwise spatial SVM trained to predict the predicate based on the four-dimensional feature of Eq. (\ref{eq:spatial_feature}). This yields an 11-dim. feature vector.
By contrast with phrase localization, our features for VRD are dense (always available for every relationship). 

In Section \ref{sec:combiningWts} we found feature weights by maximizing our recall metric. Here we have a more conventional detection task, so we obtain better performance by training a linear rank-SVM model~\cite{joachims2006training} to enforce that correctly detected relationships are ranked higher than negative detections (where either box has $<$ 0.5 IOU with the ground truth). We use the test set object detections (just the boxes, not the scores) provided by~\cite{lu2016visual} to directly compare performance with the same candidate regions. During testing, we produce a score for every ordered pair of detected boxes and all possible predicates, and retain the top 10 predicted relationships per pair of (subject, object) boxes.

Consistent with~\cite{lu2016visual}, Table~\ref{tab:vrd_results} reports recall, R@\{100, 50\}, or the portion of correctly localized relationships in the top 100 (resp. 50) ranked relationships in the image. The right side shows performance for relationships that have not been encountered in the training set. Our method clearly outperforms that of Lu~\etal~\cite{lu2016visual}, which uses separate visual, language, and relationship likelihood cues. We also outperform Zhang~\etal~\cite{zhang2017VRDTrans}, which combines object detectors, visual appearance, and object position in a single neural network.  We observe that cues based on object class and relative subject-object position provide a noticeable boost in performance. Further, due to using CCA with multi-modal embeddings, we generalize better to unseen relationships.  Qualitative examples and associated discussion can be found in Appendix~\ref{app:vrd}.

\begin{table*}
  \centering
    \begin{tabular}{|ll|cc|cc|cc|cc|}
      \hline
      & \multirow{2}{*}{Method} & \multicolumn{2}{c|}{Phrase Det.}  & \multicolumn{2}{c|}{Rel. Det.} & \multicolumn{2}{c|}{Zero-shot Phrase Det.} & \multicolumn{2}{c|}{Zero-shot Rel. Det.} \\
      & & R@100 & R@50 & R@100 & R@50 & \ \ R@100 & R@50 & \ \ R@100 & R@50 \\
      \hline \hline
      (a) & Visual Only Model~\cite{lu2016visual} & 2.61 & 2.24 & 1.85 & 1.58 & 1.12 & 0.95 & 0.78 & 0.67\\
      & Visual + Language + & \multirow{2}{*}{17.03} & \multirow{2}{*}{16.17} & \multirow{2}{*}{14.70} & \multirow{2}{*}{13.86} & \multirow{2}{*}{3.75} & \multirow{2}{*}{3.36} & \multirow{2}{*}{3.52} & \multirow{2}{*}{3.13}\\
      & Likelihood Model~\cite{lu2016visual} & & & & & & & & \\
      & VTransE~\cite{zhang2017VRDTrans} & {\bf 22.42} & {\bf 19.42} & 15.20 & 14.07 & 3.51 & 2.65 & 2.14 & 1.71\\
      \hline
      (b) & CCA & 15.36 & 11.38 & 13.69 & 10.08 & 12.40 & 7.78 & 11.12 & 6.59\\
      & CCA + Size & 15.85 & 11.72 & 14.05 & 10.36 & 12.92 & 8.04 & 11.46 & 6.76\\
      & CCA + Size + Position & 20.70 & 16.89 & {\bf 18.37} & {\bf 15.08} & {\bf 15.23} & {\bf 10.86} & {\bf 13.43} & {\bf 9.67}\\
      \hline
    \end{tabular}
       \vspace{-2mm}
\caption{Relationship detection recall at different thresholds (R@\{100,50\}). CCA refers to the combination of six CCA models (see text). Position refers to the combination of individual box position and pairwise spatial classifiers. This comparison excludes concurrent work that was published after our initial submission~\cite{liCVPR2017,liangCVPR2017}.}
\label{tab:vrd_results}
\end{table*}

\section{Conclusion}

This paper introduced a framework incorporating a comprehensive collection of image- and language-based cues for visual grounding and demonstrated significant gains over the state of the art on two tasks: phrase localization on Flickr30k Entities and relationship detection on the VRD dataset. For the latter task, we got particularly pronounced gains for the zero-shot learning scenario. In future work, we would like to train a single network for combining multiple cues.  Doing this in a unified end-to-end fashion is challenging, since one needs to find the right balance between parameter sharing and specialization or fine-tuning required by individual cues. To this end, our work provides a strong baseline and can help to inform future approaches.
\smallskip

\noindent {\bf Acknowledgments.} This work was partially supported by NSF grants 1053856, 1205627, 1405883, 1302438, and 1563727, Xerox UAC, the Sloan Foundation, and a Google Research Award.

{\small
\bibliographystyle{ieee}
\bibliography{egbib}
}

\appendix
\onecolumn
\section{Visualization of detected relationships (VRD Dataset)}
\label{app:vrd}
\begin{figure*}[h!]
  \centering
  \includegraphics[width=0.9\textwidth]{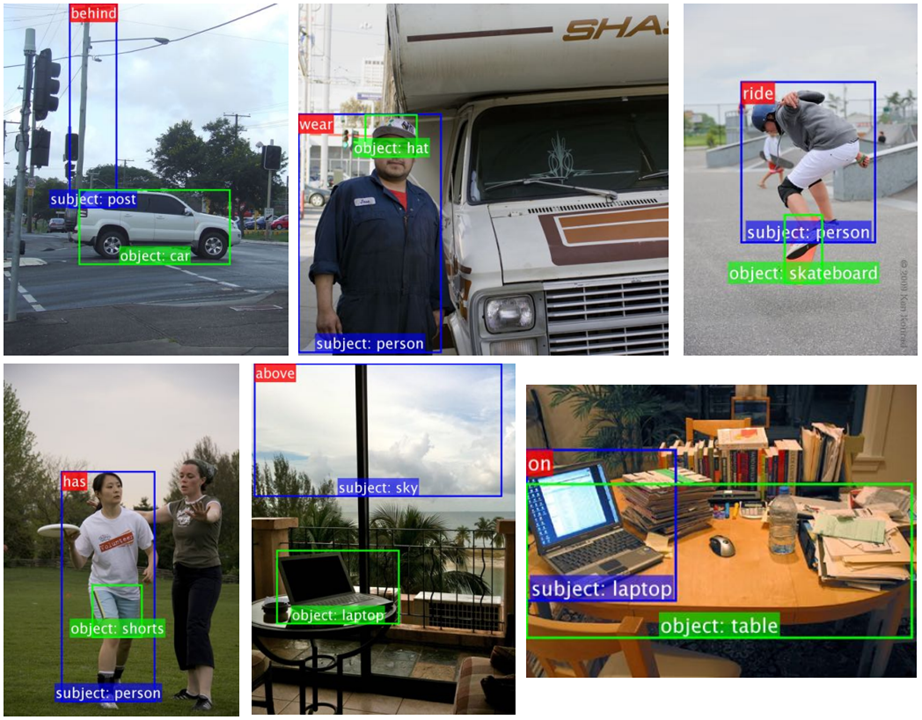}
  \caption{Highly confident and correctly localized relationships on the VRD 
  dataset.}
  \label{fig:correct}
\end{figure*}

Below are some example detections on the VRD test set. Figure~\ref{fig:correct} 
shows some of the highly confident and correctly localized detections. 
We detect different types of relationships - spatial 
(\emph{post, behind, car}), (\emph{sky, above, laptop}), (\emph{laptop, on, table}), 
clothing (\emph{person, wear, hat}), (\emph{person, has, shorts}), and actions 
(\emph{person, ride, skateboard}).

\begin{figure*}[h!]
  \centering
  \includegraphics[width=0.9\textwidth]{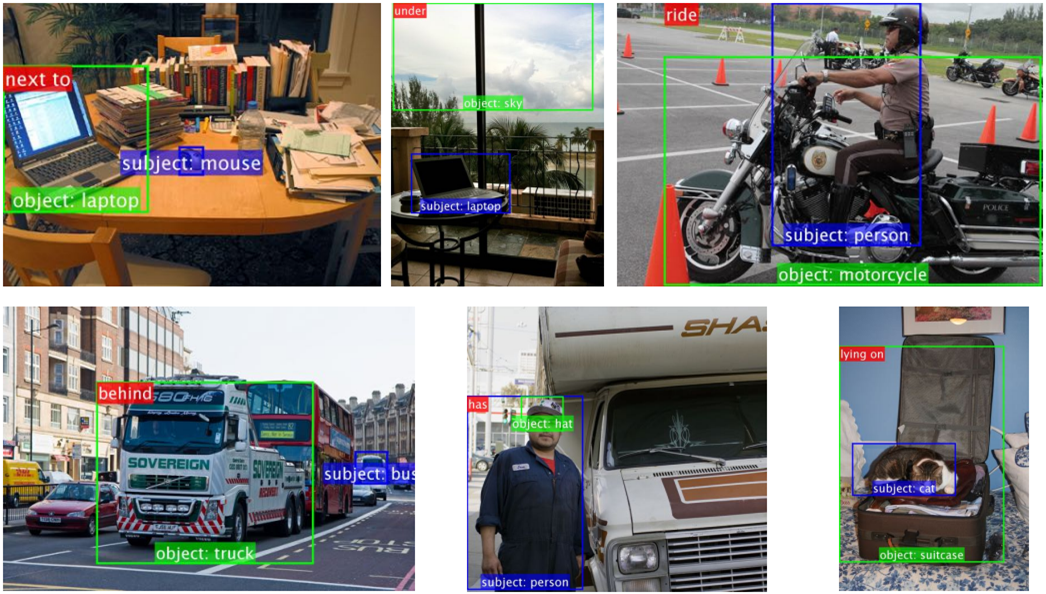}
  \caption{Plausible and logically correct detected relationships, penalized as 
  negatives due to lack of annotations in the VRD dataset.}
  \label{fig:false-neg}
\end{figure*}

Figure~\ref{fig:false-neg} shows detections which were marked as negatives by 
the evaluation code as these relationships were not annotated in the 
corresponding images. However, note that these predictions are logically correct.
The \emph{mouse} is indeed \emph{next to} the \emph{laptop} (leftmost, first row), 
and the \emph{laptop} is \emph{under} the \emph{sky} (middle, first row). 
Further, in the leftmost, second row image of Figure~\ref{fig:correct}, the 
relationship (\emph{person, has, shorts}) was marked as present, whereas the 
middle, second row image in Figure~\ref{fig:false-neg} has 
(\emph{person, has, hat}) marked as absent, which indicates a lapse in annotation.

\begin{figure*}[!h]
  \centering
  \includegraphics[width=0.9\textwidth]{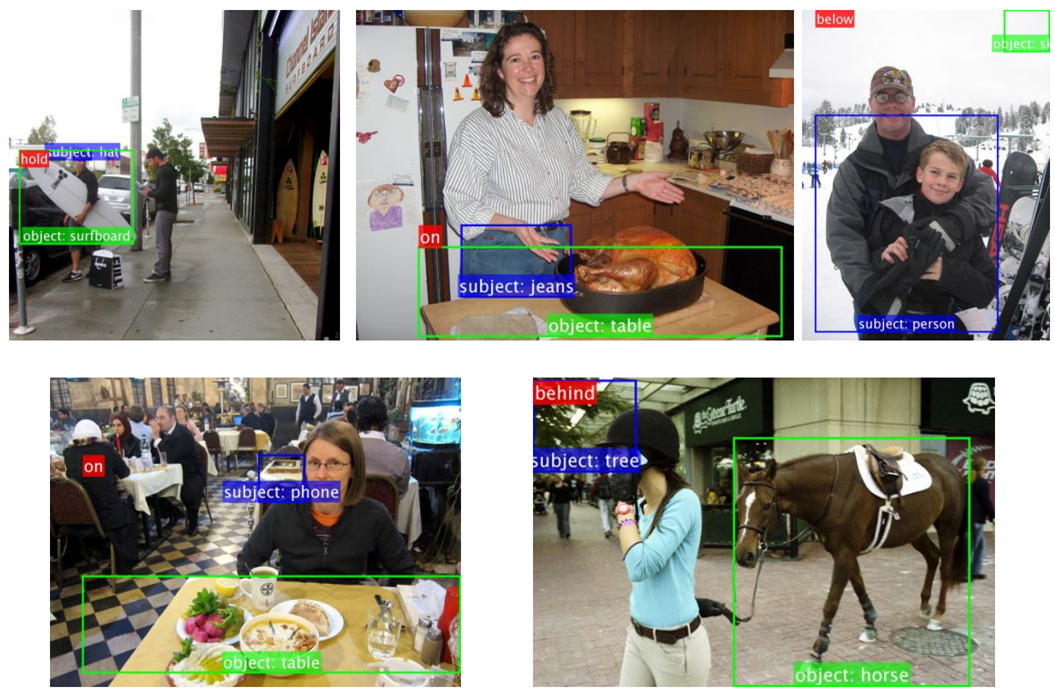}
  \caption{Falsely detected relationships on the VRD dataset. Mistakes are 
  either due to incorrect localization of objects, prediction of implausible 
  relationships, contextually incorrect relationships, or a combination of mistakes.}
  \label{fig:true-neg}
\end{figure*}

Figure~\ref{fig:true-neg} shows examples of wrongly detected relationships. 
Some of these relationships are logically implausible such as 
(\emph{hat, hold, surfboard}) (leftmost, first row), while others such as 
(\emph{jeans, on, table}) (middle, first row), while plausible, aren't 
contextually true in the image. Other failure modes include incorrect detections 
such as the \emph{sky} in the (rightmost, first row) image and the \emph{phone} 
in the (leftmost, second row) image.\\

\clearpage

\section{List of detector classes from Flickr30k Entities}
\label{app:detector}
\subsection{Adjectives}
\setcounter{rownumbers}{0}
\begin{table}[ht]
\vspace{-0.5cm}
\setlength{\tabcolsep}{3pt}
\footnotesize\begin{tabular}{llllllllllllll}
  \rownumber)&  white & \rownumber)&  people-white  & \rownumber)&  female  & \rownumber)&  empty & \rownumber)&  new &
  \rownumber)&  black & \rownumber)&  people-black  \\
   \rownumber)&  grassy  & \rownumber)&  wet & \rownumber)&  colored &
  \rownumber)&  red & \rownumber)&  people-red  & \rownumber)&  sunny & \rownumber)&  smiling\\
   \rownumber)&  professional  & \rownumber)&  brown & \rownumber)&  people-blond  & \rownumber)&  snowy & \rownumber)&  african & \rownumber)&  indoor  & \rownumber)&  gray  \\ 
   \rownumber)&  people-blue & \rownumber)&  male  & \rownumber)&  indian  & \rownumber)&  oriental &
  \rownumber)&  blond & \rownumber)&  people-green  & \rownumber)&  crowded \\
  \rownumber)&  bald  & \rownumber)&  cold & \rownumber)&  blue  & \rownumber)&  people-yellow & \rownumber)&  shirtless & \rownumber)&  american  & \rownumber)&  hot \\
  \rownumber)&  green & \rownumber)&  young & \rownumber)&  dirt  & \rownumber)&  dark-haired & \rownumber)&  dark-skinned & \rownumber)&  orange  & \rownumber)&  younger \\
  \rownumber)&  paved & \rownumber)&  teenage & \rownumber)&  cloudy & \rownumber)&  pink  & \rownumber)&  older & \rownumber)&  rocky & \rownumber)&  urban \\
  \rownumber)&  military & \rownumber)&  purple  & \rownumber)&  asian & \rownumber)&  hard  & \rownumber)&  light & \rownumber)&  hooded & \rownumber)&  yellow  \\ 
  \rownumber)&  dark  & \rownumber)&  beautiful & \rownumber)&  sandy & \rownumber)&  adult & \rownumber)&  golden  & \rownumber)&  elderly & \rownumber)&  bright  \\
  \rownumber)&  chinese & \rownumber)&  little & \rownumber)&  tan & \rownumber) &  old & \rownumber)&  concrete  & \rownumber)&  outdoors  & \rownumber)&  long  \\
  \rownumber)&  colorful  & \rownumber)&  wooden  & \rownumber)&  full  & \rownumber)&  plastic & \rownumber)&  tall  &
  \rownumber)&  striped & \rownumber)&  middle-aged \\
  \rownumber)&  multicolored  & \rownumber)&  bearded & \rownumber)&  huge  &
  \rownumber)&  short & \rownumber)&  high  & \rownumber)&  top & &    \\   
\end{tabular}
\end{table}

\subsection{Subject-Verb}
\setcounter{rownumbers}{0}
\begin{table}[H]
\vspace{-0.5cm}
\setlength{\tabcolsep}{3pt}
\footnotesize\begin{tabular}{llllllllll}
 \rownumber) & animals-catching & \rownumber) & animals-climbing & \rownumber) & animals-digging & \rownumber) & animals-fighting & \rownumber) & animals-flying\\
\rownumber)& animals-holding &\rownumber)& animals-jumping &\rownumber)& animals-playing & \rownumber)&animals-running &\rownumber)& animals-sitting\\
\rownumber)& animals-sleeping &\rownumber)& animals-splashing & \rownumber)&animals-standing &\rownumber)& animals-swimming &\rownumber)& animals-walking \\
\rownumber)& bodyparts-holding & \rownumber)& bodyparts-sitting &\rownumber)& bodyparts-walking &\rownumber)& clothing-climbing &\rownumber)& clothing-dancing \\
\rownumber)&clothing-eating &\rownumber)& clothing-holding &\rownumber)& clothing-jumping &\rownumber)& clothing-performing &\rownumber)&clothing-playing \\
\rownumber)& clothing-posing &\rownumber)& clothing-reading &\rownumber)& clothing-riding & \rownumber)&clothing-running &\rownumber)& clothing-singing \\
\rownumber)& clothing-sitting &\rownumber)& clothing-sleeping & \rownumber)&clothing-smiling &\rownumber)& clothing-standing &\rownumber)& clothing-talking \\
\rownumber)& clothing-walking &\rownumber)&clothing-working &\rownumber)& instruments-singing &\rownumber)& other-cooking &\rownumber)& other-drinking \\
\rownumber)&other-eating &\rownumber)& other-flying &\rownumber)& other-holding &\rownumber)& other-jumping &
\rownumber)&other-performing \\
\rownumber)& other-playing &\rownumber)& other-pointing &\rownumber)& other-posing & \rownumber)&other-reading &\rownumber)& other-riding \\
\rownumber)& other-running &\rownumber)& other-singing & \rownumber)&other-sitting &\rownumber)& other-sleeping &\rownumber)& other-smiling \\
\rownumber)& other-standing & \rownumber)&other-talking &\rownumber)& other-throwing &\rownumber)& other-walking &\rownumber)& other-working \\
\rownumber)&other-writing &\rownumber)& people-blowing &\rownumber)& people-catching &\rownumber)& people-cleaning &
\rownumber)&people-climbing\\
\rownumber)& people-cooking &\rownumber)& people-cutting &\rownumber)& people-dancing &
\rownumber)&people-digging &\rownumber)& people-drawing \\
\rownumber)& people-drinking &\rownumber)& people-driving &
\rownumber)&people-eating &\rownumber)& people-falling &\rownumber)& people-fighting\\
\rownumber)& people-fishing & \rownumber)&people-flying &\rownumber)& people-hiking &\rownumber)& people-hit &\rownumber)& people-holding \\
\rownumber)&people-hugging & \rownumber)& people-juggling &\rownumber)& people-jumping &\rownumber)& people-kicking &
\rownumber)&people-kissing \\
\rownumber)& people-kneeling &\rownumber)& people-laughing &\rownumber)& people-painting & \rownumber)&people-performing &\rownumber)& people-playing \\
\rownumber)& people-pointing &\rownumber)& people-posing & \rownumber)&people-pushing &\rownumber)& people-reaching &\rownumber)& people-reading \\
\rownumber)& people-riding & \rownumber)&people-running &\rownumber)& people-serving &\rownumber)& people-shopping &\rownumber)& people-singing \\
\rownumber)&people-sitting &\rownumber)& people-skiing &\rownumber)& people-sleeping &\rownumber)& people-sliding &
\rownumber)&people-smiling \\
\rownumber)& people-smoking &\rownumber)& people-splashing &\rownumber)& people-standing & \rownumber)&people-surfing &\rownumber)& people-sweeping \\
\rownumber)& people-swimming &\rownumber)& people-swinging & \rownumber)&people-talking &\rownumber)& people-throwing &\rownumber)& people-touches \\
\rownumber)& people-walking & \rownumber)&people-waving &\rownumber)& people-working &\rownumber)& people-writing &\rownumber)& scene-eating \\
\rownumber)&scene-holding &\rownumber)& scene-playing &\rownumber)& scene-reading &\rownumber)& scene-running &
\rownumber)&scene-sitting \\
\rownumber)& scene-standing &\rownumber)& scene-talking &\rownumber)& scene-walking &
\rownumber)&vehicles-driving &\rownumber)& vehicles-holding \\
\rownumber)& vehicles-running &\rownumber)& vehicles-sitting &
\rownumber)&vehicles-throwing &\rownumber)& sitting &\rownumber)& holding\\ 
\rownumber)& playing &\rownumber)&standing &\rownumber)& walking &\rownumber)& running &\rownumber)& riding \\
\rownumber)&jumping &\rownumber)& working &\rownumber)& talking &\rownumber)& performing & \rownumber)&eating \\
\rownumber)& posing &\rownumber)& climbing &\rownumber)& hiking & \rownumber)&reading &\rownumber)& dancing \\
\rownumber)& smiling &\rownumber)& singing &
\rownumber)&sleeping &\rownumber)& pushing &\rownumber)& swimming \\
\rownumber)& throwing & \rownumber)&painting &\rownumber)& driving &\rownumber)& cooking &\rownumber)& cutting \\
\rownumber)&cleaning &\rownumber)& serving &\rownumber)& swinging &\rownumber)& laughing &
\rownumber)&kicking\\
\rownumber)& hit &\rownumber)& fighting &\rownumber)& juggling &
\rownumber)&flying &\rownumber)& kissing \\
\rownumber)& pointing &\rownumber)& blowing& 
\rownumber)&sliding &\rownumber)& drinking &\rownumber)& fishing\\ 
\rownumber)& writing &
\rownumber)&skiing &\rownumber)& catching &\rownumber)& kneeling &\rownumber)& hugging \\
\rownumber)&digging &\rownumber)& smoking &\rownumber)& shopping &\rownumber)& surfing &
\rownumber)&waving \\
\rownumber)& sweeping &\rownumber)& falling &\rownumber)& reaching &
\rownumber)&drawing &\rownumber)& splashing \\
\rownumber)& touches \\
\end{tabular}
\end{table}

\subsection{Verb-Object}
\begin{table}[H]
\setcounter{rownumbers}{0}
\vspace{-1cm}
\small\begin{tabular}{llllllllll}
\rownumber)&other-blowing &\rownumber)& other-catching &\rownumber)& scene-catching &\rownumber)& other-cleaning \\
\rownumber)&scene-cleaning &\rownumber)& bodyparts-climbing &\rownumber)& other-climbing &\rownumber)& scene-climbing \\
\rownumber)&bodyparts-cooking &\rownumber)& other-cooking &\rownumber)& bodyparts-cutting &\rownumber)& other-cutting \\
\rownumber)&clothing-dancing &\rownumber)& other-dancing &\rownumber)& people-dancing &\rownumber)& scene-dancing \\
\rownumber)&other-digging &\rownumber)& scene-digging &\rownumber)& other-drawing &\rownumber)& other-drinking \\
\rownumber)&scene-drinking &\rownumber)& other-driving &\rownumber)& scene-driving &\rownumber)& vehicles-driving \\
\rownumber)&other-eating &\rownumber)& people-eating &\rownumber)& scene-eating &\rownumber)& other-falling \\
\rownumber)&scene-falling &\rownumber)& other-fighting &\rownumber)& scene-fishing &\rownumber)& other-flying \\
\rownumber)&scene-flying &\rownumber)& scene-hiking &\rownumber)& other-hit &\rownumber)& people-hit \\
\rownumber)&animals-holding &\rownumber)& bodyparts-holding &\rownumber)& clothing-holding &\rownumber)& instruments-holding \\
\rownumber)&other-holding &\rownumber)& people-holding &\rownumber)& scene-holding &\rownumber)& vehicles-holding \\
\rownumber)&people-hugging &\rownumber)& other-juggling &\rownumber)& animals-jumping &\rownumber)& bodyparts-jumping \\
\rownumber)&other-jumping &\rownumber)& people-jumping &\rownumber)& scene-jumping &\rownumber)& vehicles-jumping \\
\rownumber)&other-kicking &\rownumber)& people-kicking &\rownumber)& people-kissing &\rownumber)& scene-kissing \\
\rownumber)&other-kneeling &\rownumber)& scene-kneeling &\rownumber)& other-laughing &\rownumber)& people-laughing \\
\rownumber)&other-painting &\rownumber)& scene-painting &\rownumber)& instruments-performing &\rownumber)& other-performing \\
\rownumber)&people-performing &\rownumber)& scene-performing &\rownumber)& animals-playing &\rownumber)& clothing-playing \\
\rownumber)&instruments-playing &\rownumber)& other-playing &\rownumber)& people-playing &\rownumber)& scene-playing \\
\rownumber)&vehicles-playing &\rownumber)& bodyparts-pointing &\rownumber)& other-pointing &\rownumber)& people-pointing \\
\rownumber)&scene-pointing &\rownumber)& bodyparts-posing &\rownumber)& clothing-posing &\rownumber)& other-posing \\
\rownumber)&people-posing &\rownumber)& scene-posing &\rownumber)& other-pushing &\rownumber)& people-pushing \\
\rownumber)&vehicles-pushing &\rownumber)& other-reaching &\rownumber)& scene-reaching &\rownumber)& other-reading \\
\rownumber)&people-reading &\rownumber)& animals-riding &\rownumber)& other-riding &\rownumber)& people-riding \\
\rownumber)&scene-riding &\rownumber)& vehicles-riding &\rownumber)& animals-running &\rownumber)& bodyparts-running \\
\rownumber)&clothing-running &\rownumber)& other-running &\rownumber)& people-running &\rownumber)& scene-running \\
\rownumber)&vehicles-running &\rownumber)& other-serving &\rownumber)& people-serving &\rownumber)& other-shopping \\
\rownumber)&instruments-singing &\rownumber)& other-singing &\rownumber)& people-singing &\rownumber)& animals-sitting \\
\rownumber)&bodyparts-sitting &\rownumber)& clothing-sitting &\rownumber)& instruments-sitting &\rownumber)& other-sitting \\
\rownumber)&people-sitting &\rownumber)& scene-sitting &\rownumber)& vehicles-sitting &\rownumber)& scene-skiing \\
\rownumber)&bodyparts-sleeping &\rownumber)& other-sleeping &\rownumber)& people-sleeping &\rownumber)& scene-sleeping \\
\rownumber)&other-sliding &\rownumber)& scene-sliding &\rownumber)& bodyparts-smiling &\rownumber)& clothing-smiling \\
\rownumber)&other-smiling &\rownumber)& people-smiling &\rownumber)& scene-smiling &\rownumber)& other-smoking \\
\rownumber)&scene-splashing &\rownumber)& animals-standing &\rownumber)& bodyparts-standing &\rownumber)& clothing-standing \\
\rownumber)&other-standing &\rownumber)& people-standing &\rownumber)& scene-standing &\rownumber)& vehicles-standing \\
\rownumber)&scene-surfing &\rownumber)& other-sweeping &\rownumber)& scene-sweeping &\rownumber)& other-swimming \\
\rownumber)&scene-swimming &\rownumber)& other-swinging &\rownumber)& clothing-talking &\rownumber)& other-talking \\
\rownumber)&people-talking &\rownumber)& scene-talking &\rownumber)& other-throwing &\rownumber)& people-throwing \\
\rownumber)&scene-throwing &\rownumber)& bodyparts-touches &\rownumber)& other-touches &\rownumber)& animals-walking \\
\rownumber)&bodyparts-walking &\rownumber)& clothing-walking &\rownumber)& other-walking &\rownumber)& people-walking \\
\rownumber)&scene-walking &\rownumber)& vehicles-walking &\rownumber)& bodyparts-waving &\rownumber)& other-waving \\
\rownumber)&people-waving &\rownumber)& clothing-working &\rownumber)& other-working &\rownumber)& people-working \\
\rownumber)&scene-working &\rownumber)& vehicles-working &\rownumber)& other-writing &\rownumber)& sitting \\
\rownumber)&holding &\rownumber)& playing &\rownumber)& standing &\rownumber)& walking \\
\rownumber)&running &\rownumber)& riding &\rownumber)& jumping &\rownumber)& working \\
\rownumber)&talking &\rownumber)& performing &\rownumber)& eating &\rownumber)& posing \\
\rownumber)&climbing &\rownumber)& hiking &\rownumber)& reading &\rownumber)& dancing \\
\rownumber)&smiling &\rownumber)& singing &\rownumber)& sleeping &\rownumber)& pushing \\
\rownumber)&swimming &\rownumber)& throwing &\rownumber)& painting &\rownumber)& driving \\
\rownumber)&cooking &\rownumber)& cutting &\rownumber)& cleaning &\rownumber)& serving \\
\rownumber)&swinging &\rownumber)& laughing &\rownumber)& kicking &\rownumber)& hit \\
\rownumber)&fighting &\rownumber)& juggling &\rownumber)& flying &\rownumber)& kissing \\
\rownumber)&pointing &\rownumber)& blowing &\rownumber)& sliding &\rownumber)& drinking \\
\rownumber)&fishing &\rownumber)& writing &\rownumber)& skiing &\rownumber)& catching \\
\rownumber)&kneeling &\rownumber)& hugging &\rownumber)& digging &\rownumber)& smoking \\
\rownumber)&shopping &\rownumber)& surfing &\rownumber)& waving &\rownumber)& sweeping \\
\rownumber)&falling &\rownumber)& reaching &\rownumber)& drawing &\rownumber)& splashing \\
\rownumber)&touches \\
\end{tabular}
\end{table}

\vspace{-50pt}
\section{List of phrase-pair relationships from Flickr30k Entities}
\label{app:spatial}
\vspace{-8pt}
\subsection{Verbs}
\vspace{-5pt}
\setcounter{rownumbers}{0}
\begin{table}[H]
\vspace{-0.4cm}
\setlength{\tabcolsep}{3pt}
\footnotesize\begin{tabular}{llllllllll}
\rownumber)&dog-catching-frisbee &\rownumber)& dog-holding-stick &\rownumber)& dog-jumping-ball &\rownumber)& dog-jumping-frisbee \\
\rownumber)&dog-jumping-hurdle &\rownumber)& dog-jumping-people &\rownumber)& dog-jumping-water &\rownumber)& dog-playing-ball \\
\rownumber)&dog-running-beach &\rownumber)& dog-running-field &\rownumber)& dog-running-grass &\rownumber)& dog-running-snow \\
\rownumber)&dog-running-water &\rownumber)& dog-swimming-water &\rownumber)& dogs-playing-grass &\rownumber)& dogs-playing-snow \\
\rownumber)&dogs-running-field &\rownumber)& dogs-running-grass &\rownumber)& people-blowing-bubbles &\rownumber)& people-catching-ball \\
\rownumber)&people-catching-wave &\rownumber)& people-cleaning-dishes &\rownumber)& people-climbing-mountain &\rownumber)& people-climbing-rock \\
\rownumber)&people-climbing-rock+wall &\rownumber)& people-climbing-rocks &\rownumber)& people-climbing-tree &\rownumber)& people-climbing-wall \\
\rownumber)&people-cooking-food &\rownumber)& people-cutting-cake &\rownumber)& people-dancing-people &\rownumber)& people-dancing-stage \\
\rownumber)&people-digging-snow &\rownumber)& people-drinking-beer &\rownumber)& people-eating-food &\rownumber)& people-eating-meal \\
\rownumber)&people-eating-table &\rownumber)& people-hit-ball &\rownumber)& people-hit-tennis+ball &\rownumber)& people-holding-ball \\
\rownumber)&people-holding-book &\rownumber)& people-holding-box &\rownumber)& people-holding-camera &\rownumber)& people-holding-cup \\
\rownumber)&people-holding-dog &\rownumber)& people-holding-drink &\rownumber)& people-holding-flag &\rownumber)& people-holding-flags \\
\rownumber)&people-holding-flowers &\rownumber)& people-holding-football &\rownumber)& people-holding-guitar &\rownumber)& people-holding-microphone \\
\rownumber)&people-holding-object &\rownumber)& people-holding-people &\rownumber)& people-holding-rope &\rownumber)& people-holding-shovel \\
\rownumber)&people-holding-sign &\rownumber)& people-holding-signs &\rownumber)& people-holding-something &\rownumber)& people-holding-stick \\
\rownumber)&people-holding-tennis+racket &\rownumber)& people-hugging-people &\rownumber)& people-jumping-ball &\rownumber)& people-jumping-bed \\
\rownumber)&people-jumping-bike &\rownumber)& people-jumping-hurdle &\rownumber)& people-jumping-people &\rownumber)& people-jumping-pool \\
\rownumber)&people-jumping-ramp &\rownumber)& people-jumping-rock &\rownumber)& people-jumping-swimming+pool &\rownumber)& people-jumping-trampoline \\
\rownumber)&people-jumping-water &\rownumber)& people-kicking-ball &\rownumber)& people-kicking-people &\rownumber)& people-kicking-soccer+ball \\
\rownumber)&people-kissing-people &\rownumber)& people-laughing-people &\rownumber)& people-painting-picture &\rownumber)& people-performing-people \\
\rownumber)&people-performing-stage &\rownumber)& people-playing-accordion &\rownumber)& people-playing-bagpipes &\rownumber)& people-playing-ball \\
\rownumber)&people-playing-basketball &\rownumber)& people-playing-beach &\rownumber)& people-playing-board+game &\rownumber)& people-playing-cello \\
\rownumber)&people-playing-dog &\rownumber)& people-playing-drum &\rownumber)& people-playing-drums &\rownumber)& people-playing-flute \\
\rownumber)&people-playing-football &\rownumber)& people-playing-fountain &\rownumber)& people-playing-frisbee &\rownumber)& people-playing-game \\
\rownumber)&people-playing-guitar &\rownumber)& people-playing-guitars &\rownumber)& people-playing-instrument &\rownumber)& people-playing-instruments \\
\rownumber)&people-playing-keyboard &\rownumber)& people-playing-people &\rownumber)& people-playing-piano &\rownumber)& people-playing-pool \\
\rownumber)&people-playing-sand &\rownumber)& people-playing-saxophone &\rownumber)& people-playing-snow &\rownumber)& people-playing-soccer \\
\rownumber)&people-playing-stage &\rownumber)& people-playing-swing &\rownumber)& people-playing-toy &\rownumber)& people-playing-toys \\
\rownumber)&people-playing-trumpet &\rownumber)& people-playing-violin &\rownumber)& people-playing-volleyball &\rownumber)& people-playing-water \\
\rownumber)&people-posing-people &\rownumber)& people-posing-picture &\rownumber)& people-pushing-cart &\rownumber)& people-pushing-people \\
\rownumber)&people-pushing-stroller &\rownumber)& people-reading-book &\rownumber)& people-reading-magazine &\rownumber)& people-reading-newspaper \\
\rownumber)&people-reading-paper &\rownumber)& people-riding-bicycle &\rownumber)& people-riding-bicycles &\rownumber)& people-riding-bike \\
\rownumber)&people-riding-bikes &\rownumber)& people-riding-bull &\rownumber)& people-riding-dirt+bike &\rownumber)& people-riding-horse \\
\rownumber)&people-riding-horses &\rownumber)& people-riding-motorbike &\rownumber)& people-riding-motorcycle &\rownumber)& people-riding-people \\
\rownumber)&people-riding-scooter &\rownumber)& people-riding-skateboard &\rownumber)& people-riding-street &\rownumber)& people-riding-surfboard \\
\rownumber)&people-riding-unicycle &\rownumber)& people-riding-wave &\rownumber)& people-running-ball &\rownumber)& people-running-beach \\
\rownumber)&people-running-field &\rownumber)& people-running-grass &\rownumber)& people-running-people &\rownumber)& people-running-road \\
\rownumber)&people-running-sidewalk &\rownumber)& people-running-street &\rownumber)& people-running-track &\rownumber)& people-running-water \\
\rownumber)&people-serving-food &\rownumber)& people-singing-guitar &\rownumber)& people-singing-microphone &\rownumber)& people-singing-people \\
\rownumber)&people-sitting-beach &\rownumber)& people-sitting-bed &\rownumber)& people-sitting-bench &\rownumber)& people-sitting-benches \\
\rownumber)&people-sitting-bike &\rownumber)& people-sitting-blanket &\rownumber)& people-sitting-boat &\rownumber)& people-sitting-building \\
\rownumber)&people-sitting-chair &\rownumber)& people-sitting-chairs &\rownumber)& people-sitting-couch &\rownumber)& people-sitting-curb \\
\rownumber)&people-sitting-desk &\rownumber)& people-sitting-dock &\rownumber)& people-sitting-floor &\rownumber)& people-sitting-grass \\
\rownumber)&people-sitting-horse &\rownumber)& people-sitting-ledge &\rownumber)& people-sitting-motorcycle &\rownumber)& people-sitting-park+bench \\
\rownumber)&people-sitting-people &\rownumber)& people-sitting-rock &\rownumber)& people-sitting-rocks &\rownumber)& people-sitting-sidewalk \\
\rownumber)&people-sitting-steps &\rownumber)& people-sitting-stool &\rownumber)& people-sitting-street &\rownumber)& people-sitting-swing \\
\rownumber)&people-sitting-table &\rownumber)& people-sitting-tables &\rownumber)& people-sitting-tree &\rownumber)& people-sitting-wall \\
\rownumber)&people-sitting-water &\rownumber)& people-sleeping-bench &\rownumber)& people-sleeping-chair &\rownumber)& people-sleeping-couch \\
\rownumber)&people-sleeping-grass &\rownumber)& people-sleeping-people &\rownumber)& people-sliding-base &\rownumber)& people-sliding-slide \\
\rownumber)&people-smiling-people &\rownumber)& people-smoking-cigarette &\rownumber)& people-standing-beach &\rownumber)& people-standing-boat \\
\rownumber)&people-standing-bridge &\rownumber)& people-standing-building &\rownumber)& people-standing-car &\rownumber)& people-standing-counter \\
\rownumber)&people-standing-door &\rownumber)& people-standing-doorway &\rownumber)& people-standing-fence &\rownumber)& people-standing-field \\
\rownumber)&people-standing-grass &\rownumber)& people-standing-ladder &\rownumber)& people-standing-line &\rownumber)& people-standing-people \\
\rownumber)&people-standing-platform &\rownumber)& people-standing-podium &\rownumber)& people-standing-road &\rownumber)& people-standing-rock \\
\rownumber)&people-standing-rocks &\rownumber)& people-standing-sidewalk &\rownumber)& people-standing-sign &\rownumber)& people-standing-snow \\
\rownumber)&people-standing-stage &\rownumber)& people-standing-street &\rownumber)& people-standing-table &\rownumber)& people-standing-tree \\
\rownumber)&people-standing-wall &\rownumber)& people-standing-water &\rownumber)& people-surfing-wave &\rownumber)& people-swimming-pool \\
\rownumber)&people-swinging-bat &\rownumber)& people-swinging-swing &\rownumber)& people-talking-cellphone &\rownumber)& people-talking-microphone \\
\rownumber)&people-talking-people &\rownumber)& people-talking-phone &\rownumber)& people-throwing-ball &\rownumber)& people-throwing-frisbee \\
\rownumber)&people-throwing-people &\rownumber)& people-walking-beach &\rownumber)& people-walking-bicycle &\rownumber)& people-walking-bike \\
\rownumber)&people-walking-bridge &\rownumber)& people-walking-building &\rownumber)& people-walking-city+street &\rownumber)& people-walking-dog \\
\rownumber)&people-walking-dogs &\rownumber)& people-walking-field &\rownumber)& people-walking-grass &\rownumber)& people-walking-hill \\
\rownumber)&people-walking-path &\rownumber)& people-walking-people &\rownumber)& people-walking-road &\rownumber)& people-walking-sidewalk \\
\rownumber)&people-walking-snow &\rownumber)& people-walking-stairs &\rownumber)& people-walking-street &\rownumber)& people-walking-trail \\
\rownumber)&people-walking-wall &\rownumber)& people-walking-water &\rownumber)& people-working-machine &\rownumber)& people-working-people \\
\end{tabular}
\end{table}
\vspace{-3cm}

\subsection{Prepositions}
\begin{table}[H]
\setcounter{rownumbers}{0}
\setlength{\tabcolsep}{6pt}
\vspace{-5cm}
\small\begin{tabular}{llllllllll}
\rownumber)&ball-in-mouth &\rownumber)& bicycle-on-street &\rownumber)& boat-in-water &\rownumber)& building-in-people \\
\rownumber)&dog-in-ball &\rownumber)& dog-in-collar &\rownumber)& dog-in-dog &\rownumber)& dog-in-field \\
\rownumber)&dog-in-grass &\rownumber)& dog-in-snow &\rownumber)& dog-in-stick &\rownumber)& dog-in-toy \\
\rownumber)&dog-in-water &\rownumber)& dog-on-beach &\rownumber)& dog-on-grass &\rownumber)& dog-on-hind+legs \\
\rownumber)&dog-on-leash &\rownumber)& dogs-in-dogs &\rownumber)& dogs-in-field &\rownumber)& dogs-in-grass \\
\rownumber)&dogs-in-snow &\rownumber)& dogs-in-water &\rownumber)& dogs-on-grass &\rownumber)& guitar-in-people \\
\rownumber)&hands-in-people &\rownumber)& object-in-mouth &\rownumber)& one-in-shirt &\rownumber)& other-in-shirt \\
\rownumber)&people-across-street &\rownumber)& people-behind-building &\rownumber)& people-behind-counter &\rownumber)& people-behind-fence \\
\rownumber)&people-behind-people &\rownumber)& people-between-people &\rownumber)& people-in-area &\rownumber)& people-in-back \\
\rownumber)&people-in-ball &\rownumber)& people-in-bed &\rownumber)& people-in-bicycle &\rownumber)& people-in-bike \\
\rownumber)&people-in-blanket &\rownumber)& people-in-boat &\rownumber)& people-in-body+water &\rownumber)& people-in-building \\
\rownumber)&people-in-camera &\rownumber)& people-in-cane &\rownumber)& people-in-canoe &\rownumber)& people-in-car \\
\rownumber)&people-in-cart &\rownumber)& people-in-chair &\rownumber)& people-in-chairs &\rownumber)& people-in-cigarette \\
\rownumber)&people-in-colors &\rownumber)& people-in-dirt &\rownumber)& people-in-dog &\rownumber)& people-in-dogs \\
\rownumber)&people-in-doorway &\rownumber)& people-in-face+paint &\rownumber)& people-in-field &\rownumber)& people-in-flowers \\
\rownumber)&people-in-football &\rownumber)& people-in-fountain &\rownumber)& people-in-gear &\rownumber)& people-in-grass \\
\rownumber)&people-in-guitar &\rownumber)& people-in-highchair &\rownumber)& people-in-instruments &\rownumber)& people-in-kayak \\
\rownumber)&people-in-kitchen &\rownumber)& people-in-lake &\rownumber)& people-in-line &\rownumber)& people-in-microphone \\
\rownumber)&people-in-mirror &\rownumber)& people-in-mud &\rownumber)& people-in-number &\rownumber)& people-in-ocean \\
\rownumber)&people-in-park &\rownumber)& people-in-people &\rownumber)& people-in-pool &\rownumber)& people-in-river \\
\rownumber)&people-in-room &\rownumber)& people-in-sand &\rownumber)& people-in-snow &\rownumber)& people-in-soccer+ball \\
\rownumber)&people-in-street &\rownumber)& people-in-stroller &\rownumber)& people-in-swimming+pool &\rownumber)& people-in-swing \\
\rownumber)&people-in-towel &\rownumber)& people-in-toy &\rownumber)& people-in-toys &\rownumber)& people-in-tree \\
\rownumber)&people-in-tub &\rownumber)& people-in-water &\rownumber)& people-in-wheelchair &\rownumber)& people-in-yard \\
\rownumber)&people-near-beach &\rownumber)& people-near-brick+wall &\rownumber)& people-near-building &\rownumber)& people-near-car \\
\rownumber)&people-near-fence &\rownumber)& people-near-fountain &\rownumber)& people-near-lake &\rownumber)& people-near-people \\
\rownumber)&people-near-pole &\rownumber)& people-near-road &\rownumber)& people-near-sidewalk &\rownumber)& people-near-street \\
\rownumber)&people-near-table &\rownumber)& people-near-tree &\rownumber)& people-near-wall &\rownumber)& people-near-water \\
\rownumber)&people-near-window &\rownumber)& people-on-back &\rownumber)& people-on-balcony &\rownumber)& people-on-beach \\
\rownumber)&people-on-bed &\rownumber)& people-on-bench &\rownumber)& people-on-benches &\rownumber)& people-on-bicycle \\
\rownumber)&people-on-bicycles &\rownumber)& people-on-bike &\rownumber)& people-on-bikes &\rownumber)& people-on-blanket \\
\rownumber)&people-on-board &\rownumber)& people-on-boat &\rownumber)& people-on-bridge &\rownumber)& people-on-building \\
\rownumber)&people-on-bus &\rownumber)& people-on-cellphone &\rownumber)& people-on-chair &\rownumber)& people-on-chairs \\
\rownumber)&people-on-city+street &\rownumber)& people-on-cliff &\rownumber)& people-on-computer &\rownumber)& people-on-couch \\
\rownumber)&people-on-curb &\rownumber)& people-on-deck &\rownumber)& people-on-dock &\rownumber)& people-on-fence \\
\rownumber)&people-on-field &\rownumber)& people-on-floor &\rownumber)& people-on-grass &\rownumber)& people-on-grill \\
\rownumber)&people-on-hill &\rownumber)& people-on-horse &\rownumber)& people-on-horses &\rownumber)& people-on-ice \\
\rownumber)&people-on-ladder &\rownumber)& people-on-lawn &\rownumber)& people-on-ledge &\rownumber)& people-on-machine \\
\rownumber)&people-on-mat &\rownumber)& people-on-motorcycle &\rownumber)& people-on-motorcycles &\rownumber)& people-on-mountain \\
\rownumber)&people-on-park+bench &\rownumber)& people-on-path &\rownumber)& people-on-pavement &\rownumber)& people-on-people \\
\rownumber)&people-on-phone &\rownumber)& people-on-pier &\rownumber)& people-on-platform &\rownumber)& people-on-porch \\
\rownumber)&people-on-raft &\rownumber)& people-on-rail &\rownumber)& people-on-railing &\rownumber)& people-on-ramp \\
\rownumber)&people-on-road &\rownumber)& people-on-rock &\rownumber)& people-on-rocks &\rownumber)& people-on-roof \\
\rownumber)&people-on-rope &\rownumber)& people-on-sand &\rownumber)& people-on-scaffold &\rownumber)& people-on-scaffolding \\
\rownumber)&people-on-scooter &\rownumber)& people-on-shore &\rownumber)& people-on-side+road &\rownumber)& people-on-sidewalk \\
\rownumber)&people-on-skateboard &\rownumber)& people-on-sled &\rownumber)& people-on-slide &\rownumber)& people-on-snowboard \\
\rownumber)&people-on-soccer+field &\rownumber)& people-on-sofa &\rownumber)& people-on-stage &\rownumber)& people-on-stairs \\
\rownumber)&people-on-step &\rownumber)& people-on-steps &\rownumber)& people-on-stilts &\rownumber)& people-on-stool \\
\rownumber)&people-on-street &\rownumber)& people-on-surfboard &\rownumber)& people-on-swing &\rownumber)& people-on-table \\
\rownumber)&people-on-tire+swing &\rownumber)& people-on-track &\rownumber)& people-on-trail &\rownumber)& people-on-train \\
\rownumber)&people-on-trampoline &\rownumber)& people-on-tree &\rownumber)& people-on-walkway &\rownumber)& people-on-wall \\
\rownumber)&people-on-water &\rownumber)& people-on-wave &\rownumber)& people-under-tree &\rownumber)& shirt-in-people \\
\rownumber)&something-in-mouth &\rownumber)& stick-in-mouth &\rownumber)& street-in-people &\rownumber)& table-in-people \\
\rownumber)&tattoo-on-people &\rownumber)& tennis+ball-in-mouth &\rownumber)& toy-in-mouth &\rownumber)& wall-in-graffiti \\
\end{tabular}
\end{table}

\subsection{Clothing and Body Part Attachment}
\setcounter{rownumbers}{0}
\setlength{\tabcolsep}{6pt}
\footnotesize\begin{tabular}{llllllllll}
\rownumber)&people-apron &\rownumber)& people-aprons &\rownumber)& people-arms &\rownumber)& people-attire \\
\rownumber)&people-backpack &\rownumber)& people-backpacks &\rownumber)& people-bag &\rownumber)& people-bags \\
\rownumber)&people-ball+cap &\rownumber)& people-bandanna &\rownumber)& people-baseball+cap &\rownumber)& people-baseball+uniform \\
\rownumber)&people-bathing+suit &\rownumber)& people-bathing+suits &\rownumber)& people-beanie &\rownumber)& people-beard \\
\rownumber)&people-beret &\rownumber)& people-bikini &\rownumber)& people-bikinis &\rownumber)& people-black \\
\rownumber)&people-black+shirt &\rownumber)& people-black+white &\rownumber)& people-blond-hair &\rownumber)& people-blouse \\
\rownumber)&people-blue &\rownumber)& people-body &\rownumber)& people-boots &\rownumber)& people-brown \\
\rownumber)&people-brown+jacket &\rownumber)& people-brown+shirt &\rownumber)& people-business+attire &\rownumber)& people-business+suit \\
\rownumber)&people-camouflage &\rownumber)& people-cap &\rownumber)& people-checkered+shirt &\rownumber)& people-clothes \\
\rownumber)&people-clothing &\rownumber)& people-coat &\rownumber)& people-coats &\rownumber)& people-collared+shirt \\
\rownumber)&people-costume &\rownumber)& people-costumes &\rownumber)& people-cowboy+hat &\rownumber)& people-cowboy+hats \\
\rownumber)&people-curly+hair &\rownumber)& people-denim+jacket &\rownumber)& people-dreadlocks &\rownumber)& people-dress \\
\rownumber)&people-dress+shirt &\rownumber)& people-dresses &\rownumber)& people-eyes &\rownumber)& people-face \\
\rownumber)&people-faces &\rownumber)& people-feet &\rownumber)& people-finger &\rownumber)& people-fingers \\
\rownumber)&people-flip-flops &\rownumber)& people-garb &\rownumber)& people-glasses &\rownumber)& people-gloves \\
\rownumber)&people-goggles &\rownumber)& people-gold &\rownumber)& people-gray &\rownumber)& people-green \\
\rownumber)&people-hair &\rownumber)& people-haircut &\rownumber)& people-hand &\rownumber)& people-hands \\
\rownumber)&people-harness &\rownumber)& people-hat &\rownumber)& people-hats &\rownumber)& people-head \\
\rownumber)&people-headband &\rownumber)& people-headphones &\rownumber)& people-heads &\rownumber)& people-headscarf \\
\rownumber)&people-heels &\rownumber)& people-helmet &\rownumber)& people-helmets &\rownumber)& people-hoodie \\
\rownumber)&people-jacket &\rownumber)& people-jackets &\rownumber)& people-jean+shorts &\rownumber)& people-jeans \\
\rownumber)&people-jersey &\rownumber)& people-jerseys &\rownumber)& people-jumpsuit &\rownumber)& people-khaki+pants \\
\rownumber)&people-kilt &\rownumber)& people-knees &\rownumber)& people-lab+coat &\rownumber)& people-lap \\
\rownumber)&people-leather+jacket &\rownumber)& people-leg &\rownumber)& people-legs &\rownumber)& people-leotard \\
\rownumber)&people-life+jacket &\rownumber)& people-life+jackets &\rownumber)& people-makeup &\rownumber)& people-mask \\
\rownumber)&people-mohawk &\rownumber)& people-mouth &\rownumber)& people-mustache &\rownumber)& people-necklace \\
\rownumber)&people-nose &\rownumber)& people-orange &\rownumber)& people-orange+dress &\rownumber)& people-orange+hat \\
\rownumber)&people-orange+jacket &\rownumber)& people-orange+shirt &\rownumber)& people-orange+vest &\rownumber)& people-orange+vests \\
\rownumber)&people-outfit &\rownumber)& people-outfits &\rownumber)& people-overalls &\rownumber)& people-pajamas \\
\rownumber)&people-pants &\rownumber)& people-people &\rownumber)& people-pigtails &\rownumber)& people-pink \\
\rownumber)&people-pink+coat &\rownumber)& people-pink+dress &\rownumber)& people-pink+hat &\rownumber)& people-pink+jacket \\
\rownumber)&people-pink+outfit &\rownumber)& people-pink+pants &\rownumber)& people-pink+shirt &\rownumber)& people-pink+sweater \\
\rownumber)&people-plaid+shirt &\rownumber)& people-polo+shirt &\rownumber)& people-ponytail &\rownumber)& people-purple \\
\rownumber)&people-purple+shirt &\rownumber)& people-purse &\rownumber)& people-red &\rownumber)& people-red+white \\
\rownumber)&people-red-hair &\rownumber)& people-ring &\rownumber)& people-robe &\rownumber)& people-robes \\
\rownumber)&people-rock+face &\rownumber)& people-safety+vest &\rownumber)& people-safety+vests &\rownumber)& people-sandals \\
\rownumber)&people-scarf &\rownumber)& people-scrubs &\rownumber)& people-shirt &\rownumber)& people-shirts \\
\rownumber)&people-shoe &\rownumber)& people-shoes &\rownumber)& people-shopping+bag &\rownumber)& people-shopping+bags \\
\rownumber)&people-shorts &\rownumber)& people-shoulder &\rownumber)& people-shoulders &\rownumber)& people-skirt \\
\rownumber)&people-skirts &\rownumber)& people-sleeveless+shirt &\rownumber)& people-smile &\rownumber)& people-sneakers \\
\rownumber)&people-snowshoes &\rownumber)& people-snowsuit &\rownumber)& people-socks &\rownumber)& people-straw+hat \\
\rownumber)&people-striped+shirt &\rownumber)& people-suit &\rownumber)& people-suits &\rownumber)& people-sunglasses \\
\rownumber)&people-suspenders &\rownumber)& people-sweater &\rownumber)& people-sweatshirt &\rownumber)& people-swim+trunks \\
\rownumber)&people-swimming+trunks &\rownumber)& people-swimsuit &\rownumber)& people-swimsuits &\rownumber)& people-t-shirt \\
\rownumber)&people-t-shirts &\rownumber)& people-tan+jacket &\rownumber)& people-tan+pants &\rownumber)& people-tan+shirt \\
\rownumber)&people-tank &\rownumber)& people-tank+top &\rownumber)& people-tattoo &\rownumber)& people-tattoos \\
\rownumber)&people-teeth &\rownumber)& people-thumbs &\rownumber)& people-tie &\rownumber)& people-tongue \\
\rownumber)&people-top &\rownumber)& people-tops &\rownumber)& people-trunks &\rownumber)& people-turban \\
\rownumber)&people-tuxedo &\rownumber)& people-umbrella &\rownumber)& people-umbrellas &\rownumber)& people-underwear \\
\rownumber)&people-uniform &\rownumber)& people-uniforms &\rownumber)& people-vest &\rownumber)& people-vests \\
\rownumber)&people-wedding+dress &\rownumber)& people-wetsuit &\rownumber)& people-white &\rownumber)& people-wig \\
\rownumber)&people-winter+clothes &\rownumber)& people-winter+clothing &\rownumber)& people-yellow \\
\end{tabular}

\end{document}